\pdfoutput=1

\documentclass[11pt]{article}

\usepackage[preprint]{acl}

\usepackage{times}
\usepackage{latexsym}
\usepackage{placeins}
\usepackage{subcaption}
\usepackage{enumitem}

\usepackage[T1]{fontenc}

\usepackage[utf8x]{inputenc}

\usepackage{microtype}

\usepackage{inconsolata}

\usepackage{graphicx}

\usepackage{amsmath,amsfonts,bm}

\def\eqref#1{equation~\ref{#1}}

\def\1{\bm{1}}

\DeclareMathAlphabet{\mathsfit}{\encodingdefault}{\sfdefault}{m}{sl}
\SetMathAlphabet{\mathsfit}{bold}{\encodingdefault}{\sfdefault}{bx}{n}

\usepackage{hyperref}
\usepackage{url}
\usepackage{graphicx}
\usepackage{listings}
\usepackage{booktabs}       
\usepackage{multirow}

\DeclareUnicodeCharacter{2200}{$\forall$}
\DeclareUnicodeCharacter{65533}{$\rightarrow$}

\newtheorem{definition}{Definition}

\definecolor{backcolor}{rgb}{0.95,0.95,0.95}

\lstdefinestyle{mystyle}{
    backgroundcolor=\color{backcolor},
    breaklines=true,
    numbers=left,
    numbersep=5pt,
    basicstyle=\ttfamily\footnotesize,
    captionpos=b,
    escapeinside={\%*}{*)},
}
\title{Intermediate Languages Matter: Formal Choice Drives Neurosymbolic LLM Reasoning}
\author{$\textbf{Alexander Beiser}^1$ $\textbf{David Penz}^{1,2}$ $\textbf{Nysret Musliu}^{1}$\\
\texttt{alexander.beiser@tuwien.ac.at} \\
$^1\text{TU Wien, Vienna, Austria}$ $^2\text{Johannes Kepler University Linz, Linz, Austria}$ }

\begin{document}

\maketitle
\begin{abstract}
%
%
Large language models (LLMs) achieve astonishing results on a wide range of tasks.
However, their formal reasoning ability still lags behind.
A promising approach is Neurosymbolic LLM reasoning.
It works by using LLMs as translators from natural to formal languages
and symbolic solvers for deriving correct results.
Still, it remains unclear what the contributing factors to the success of Neurosymbolic LLM reasoning are.
%
%
%
%
%
This paper shows that one important factor is the choice of the formal language.
By comparing 4 formal languages on 3 datasets over 6 LLMs,
we show that the choice of formal language affects both the syntactic and the semantic reasoning capability.
Thereby, we introduce the intermediate language challenge,
which is the challenge of picking a suitable 
formal language for neurosymbolic reasoning.
Further, we compare the effects of using different in-context-learning examples
in an ablation study.
We conclude that on average,
context-aware encodings help LLMs to reason,
while there is no apparent effect of using comments or markdown syntax.

%
%
%
%
\end{abstract}

\section{Introduction}
\label{sec:introduction}
Logical reasoning tasks pose a challenge to Large Language Models (LLMs), 
as they struggle to reason abstractly and correctly~\cite{saparov_language_2023,lampinen_language_2024,panas_can_2024}.
This leads to their sometimes spectacular failures, like deriving that birds have four legs~\cite{lin_birds_2020}.
%
%
%
%
One attempt to improve the abstract reasoning capability is Chain of Thought (CoT)~\citep{wei_chain--thought_2022} prompting.
With CoT, LLMs are nudged to reason step-by-step.
However, LLMs' step-by-step reasoning is generally \emph{non-faithful} - even when all individual reasoning steps are correct on their own,
the final conclusion can be false~\citep{lyu_faithful_2023}.
%

\emph{Neurosymbolic LLM reasoning} enables faithful reasoning chains.
It works in two steps:
the first step translates a natural language-posed logical reasoning problem 
into a \emph{formal intermediate language}.
The translation uses the \emph{in-context-learning} (ICL) capability of LLMs.
The second step is to solve the translated problem by a symbolic reasoner.
%
State-of-the-art neurosymbolic approaches,
such as Logic-LM~\cite{pan_logic-lm_2023} and LINC~\cite{olausson_linc_2023},
report substantial improvements over pure LLM prompting.

However, it remains unclear what the reasons for their reported success are.
This comes, as there are a plethora of possible contributing factors,
ranging from the LLM training data,
over auxiliary systems (such as re-prompting on errors),
to the choice of formal language.
We investigate the choice of formal language, as it is rarely justified, let alone supported by empirical evidence, leaving its impact on neurosymbolic LLM reasoning largely uncharted.


\noindent
\textbf{Contributions}.
By measuring the impact of different formal languages,
we take a first step toward better understanding why neurosymbolic systems obtain state-of-the-art results
and how the choice of formal language affects reasoning.
The main contributions of this work are:
\vspace{-0.3em}

\begin{figure*}[t]
    \includegraphics[width=14.5cm]{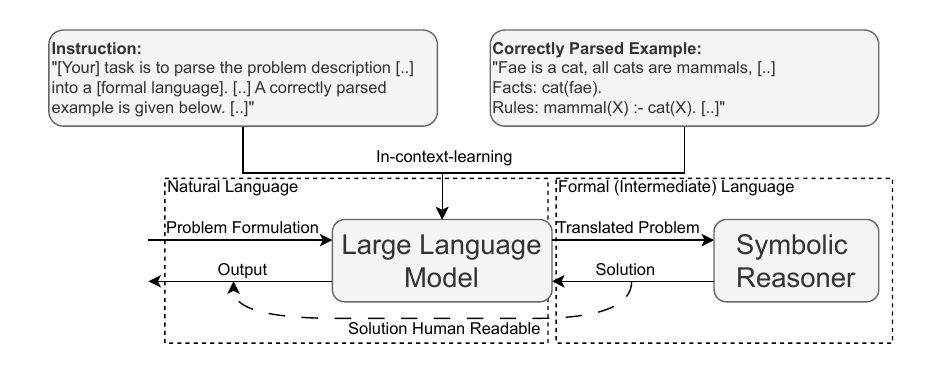}
    \vspace{-1.4em}
    \caption{
        Neurosymbolic LLM reasoning: 
        A problem formulated in natural language is translated by using in-context-learning
        into a formal language.
        Subsequently, a symbolic reasoner subsequently computes a solution to the problem, 
        which is followed by the re-translation of the solution.
    }
    \label{sec:logic_lm:neurosymbolic_schematics}
    \vspace{-1em}
\end{figure*}
\begin{itemize}[leftmargin=*]
    \item We introduce the \textit{intermediate language challenge}:
         selecting the right formal language for neurosymbolic LLM reasoning.
        \vspace{-.6em}
\item We conduct an extensive empirical study\footnote{
Our supplementary material \url{https://tinyurl.com/intermediate-language}
contains the experiment code/data.} of four formal languages across three logical reasoning datasets (ProntoQA, ProofWriter, FOLIO) and six LLMs (8B--671B).
\vspace{-0.6em}
%
\item  We perform a systematic ablation study on ICL-example encoding strategies, isolating the effects of \textit{context}, \textit{comments}, and \textit{markdown syntax}.
\vspace{-0.3em}
\end{itemize}
    Our experiments show that the choice of formal language matters:
    first-order logic outperforms logic programming languages.
    %
    Echoing earlier findings on symbolic reasoning~\cite{lampinen_language_2024},
    our neurosymbolic ablation confirms that added context improves LLM performance.
    In contrast, adding comments or Markdown markup yields no systematic benefit.
    %

We will continue after this introduction
with the preliminaries and definitions (Section~\ref{sec:background}).
Next, we introduce the intermediate language problem (Section~\ref{sec:the-intermediate-language-problem}),
which we follow with our experimental setup (Section~\ref{section:experimental-scenarios}), and
our experimental results (Section~\ref{section:experiments:experiments}).
We close our paper with the related work in Section~\ref{sec:related-work}
and our conclusions in Section~\ref{sec:discussion}.
%
%

\vspace{-0.3em}
\section{Preliminaries}
\label{sec:background}
\vspace{-0.1cm}
We briefly present the necessary background material and definitions for understanding the paper.
Recall that the main objective of this study is to compare the reasoning performance of different formal languages on modern LLMs.
By taking the perspective of end-users,
we treat LLMs as immutable black-box next-token predictor machines.
Therefore, we are mainly interested in what effects different prompting strategies have on the reasoning performance.
We consider the effects of other techniques, such as fine-tuning, as out of scope.
Throughout this paper, the terms \emph{syntax} and \emph{semantics} are used in their formal language sense.
%
%
%
%
%
%
\subsection{Chain-of-Thought (CoT) prompting}

Chain-of-Thought (CoT) prompting is an \textit{in-context-learning} (ICL)
technique with applications ranging from helping 
LLMs to express their uncertainty~\citep{xiong_can_2024},
to improving the reasoning capabilities of LLMs~\citep{wei_chain--thought_2022}.
CoT nudges the LLM to mimic a reasoning chain,
where we show an example in the next listing.
%
%
\begin{lstlisting}
The following example showcases the line of reasoning you have to follow:
---- Question ----
Each cat is a carnivore. Fae is a cat.
True or false: Fae is a carnivore
---- Reasoning ----
Fae is a cat. Each cat is a carnivore. So Fae is a carnivore.
\end{lstlisting}\vspace{-0.2cm}
%
Reasoning chains are \textit{faithful}
whenever the result follows from the individual steps in the reasoning chain.
However, LLMs' reasoning chains are \textit{non-faithful} in general~\citep{lyu_faithful_2023}.
%
%
\subsection{Neurosymbolic LLM Reasoning}
\label{sec:logic-lm}
%
Figure~\ref{sec:logic_lm:neurosymbolic_schematics} depicts the high-level schematics of neurosymbolic LLM reasoning.
%
%
%
%
A natural language-posed problem is translated into its \textit{formal language} representation by using ICL.
ICL comprises an \emph{instruction} and an \emph{example},
sometimes called \emph{ICL-instruction} and \emph{ICL-example}.
The instruction describes the general task,
while the example showcases how to translate the natural language-posed problem into a formal language.
We refer to the formal language of the ICL-example,
as the \textit{chosen} formal language.

In a second step, the symbolic reasoner solves the problem by obtaining a solution from the formal representation,
which can be either re-translated into natural language or directly used as output.
Logic-LM~\citep{pan_logic-lm_2023} and Logic-LM++~\citep{kirtania_logic-lm_2024} re-prompt the LLM in case of syntax errors, with the error message of the symbolic reasoner.
LINC~\cite{olausson_linc_2023} uses multiple ICL-examples for one problem
and prompts the LLM non-deterministically multiple times, where a majority vote is taken to get the final output.
%
%
Most approaches use a backup strategy on syntax errors, such as CoT prompting or random choice.
We abstain from a backup strategy, as we want to measure the impact of the formal language directly.

\section{The Intermediate Language Challenge for Logical Reasoning}
\label{sec:the-intermediate-language-problem}

We proceed to define our intermediate language challenge for neurosymbolic LLM reasoning.
We assume to have given a natural language-posed reasoning problem $\mathcal{P}$ and
a set of possible formal languages $\mathcal{L}$.
\begin{definition}
    The intermediate language challenge is the task of
    choosing a formal language $l \in \mathcal{L}$ for solving $\mathcal{P}$ with a high 
    reasoning accuracy.
\end{definition}
%
Inherent to the intermediate language challenge is autoformalization~\citep{wu_autoformalization_2022}.
\begin{definition}
Let $l \in \mathcal{L}$ be a fixed formal language.
Then, autoformalization aims for automatic and correct translation of $\mathcal{P}$ into $l$.
\end{definition}
While autoformalization is concerned with the correct translation from natural language into a fixed formal language $l$,
the intermediate language challenge is about choosing a suitable formal language $l' \in \mathcal{L}$ s.t. autoformalization can be done effectively.
%
%
%
We identify two root causes of the intermediate language problem:
%
%
(i) Syntax affects LLMs' reasoning performance, and (ii) one logical problem can be translated into multiple formal languages.

\noindent
\textbf{Syntax affects LLMs' reasoning performance}.
Consider the following two logical reasoning problems:
(1) ``Tommi is a tumpus. Each tumpus is a wumpus. Is Tommi a wumpus?''
(2) ``Tommi is a cat. Each cat is an animal. Is Tommi an animal?''
Recent work suggests that, on average, LLMs perform better for scenarios of type (2) than type (1)~\citep{saparov_language_2023,lampinen_language_2024}.
From a semantic perspective, both scenarios require the application of modus ponens.
Thus, as the \emph{only} difference lies in the \emph{syntax}, we can conclude that the syntax affects LLMs' reasoning capabilities.
Going back to formal languages,
observe that the syntax of formal languages differs.
Therefore, we conclude that the choice of formal language affects LLMs' reasoning capabilities.

%
%

\noindent
\textbf{Logical problems can be encoded in different formal languages}.
Take the logical reasoning problem (2) from the paragraph above. 
This problem can be encoded in different formal languages, such as logic programming or first-order logic (FOL),
while maintaining semantic correctness.
%
%
%

\section{Experiment Setup}
\label{section:experimental-scenarios}
To show the impact of the intermediate language challenge, we investigate a set of formal languages $\mathcal{L} = \{\text{Pyke},\text{ASP},\text{NLTK},\text{FOL}\}$.
We conduct experiments on three different datasets, ProntoQA~\cite{saparov_language_2023}, ProofWriter~\cite{tafjord_proofwriter_2021}, and FOLIO~\cite{han_folio_2024}.
Let $\mathcal{D}$ be a given dataset, then each data instance $\mathcal{P} \in \mathcal{D}$ can be considered a reasoning problem.
For evaluation we use six LLMs, ranging from 8B to 671B parameters.
%
%

\begin{figure*}[t]
    \begin{subfigure}{0.5\textwidth}
        \includegraphics[width=7.7cm]{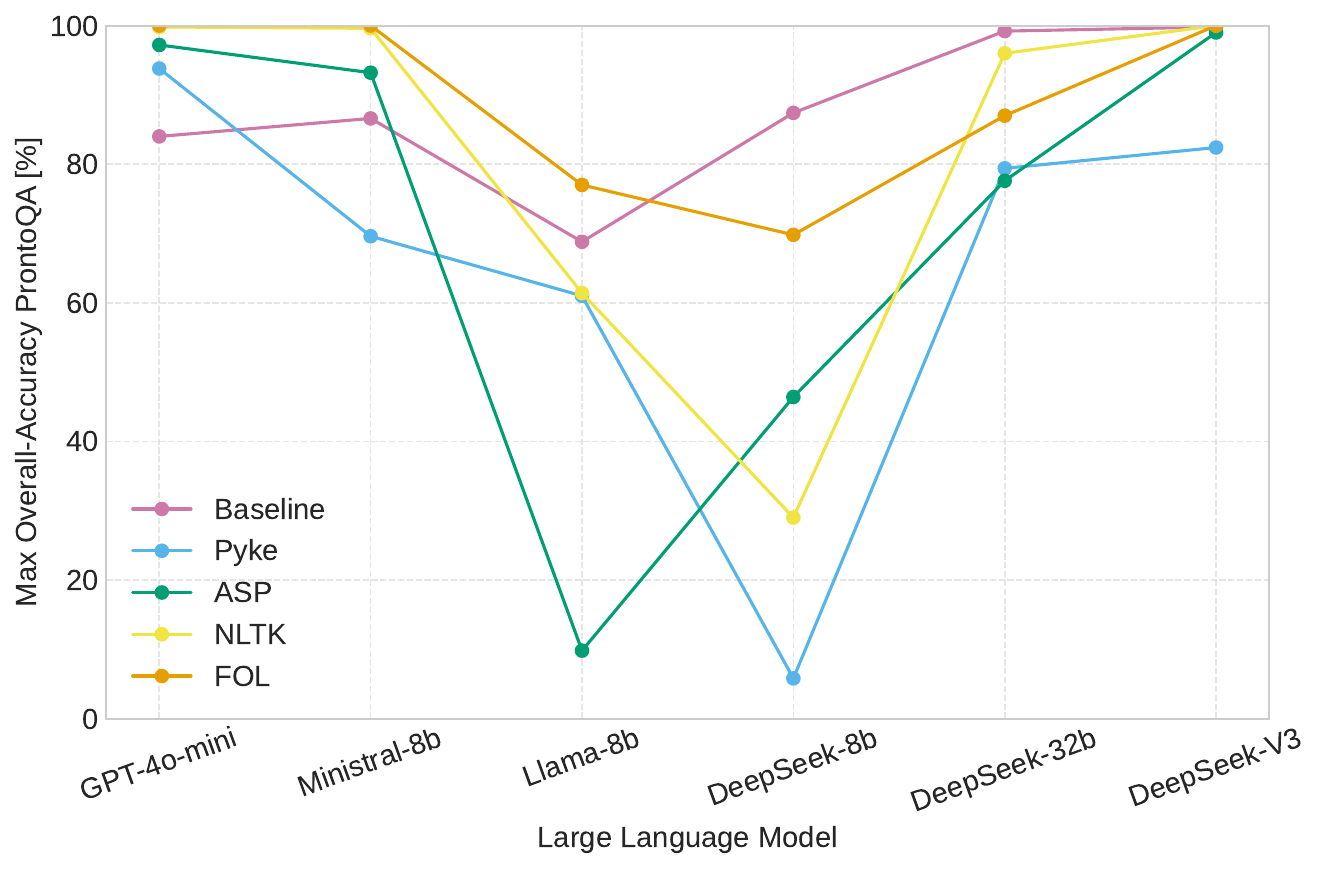}
    \end{subfigure}
    \begin{subfigure}{0.5\textwidth}
        \includegraphics[width=7.7cm]{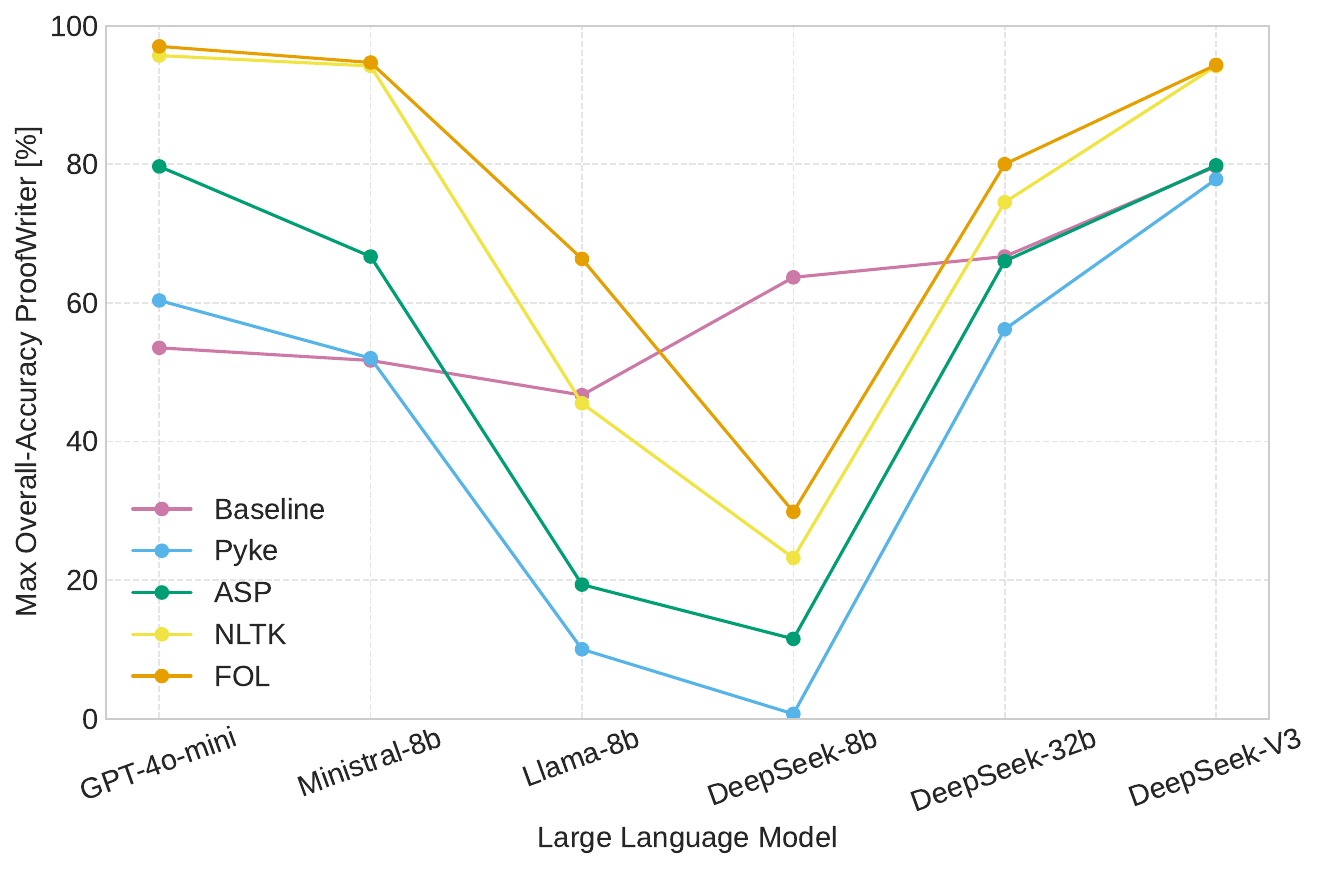}
    \end{subfigure}
    \vspace{-0.5em}
    \caption{
    Parallel coordinates plot showing maximum overall-accuracy for the ProntoQA (left) and ProofWriter (right) datasets for all LLMs.
    Maximum is computed w.r.t. all ablation study scenarios per formal language.
    }
    \vspace{-0.8em}
    \label{fig:max-results-prontoqa-proofwriter}
\end{figure*}

\subsection{Formal Languages}

We will provide a brief overview of the formal languages $\mathcal{L}$ used for our experiments.

\noindent \textbf{Pyke}:
The logic programming derivative Pyke~\citep{frederiksen_applying_2008}
expresses rules similar to \textit{if}-\textit{then} statements.
One defines a fact- and a rule-base,
where conclusions are reached by forward, or backward chaining algorithms.
%
We show a translation of problem (2) into Pyke's syntax.
%
\begin{lstlisting}
Cat(Tommi, True)
fact1
	foreach
		facts.Cat($x, True)
	assert
		facts.Animal($x, True)
\end{lstlisting}

\noindent \textbf{ASP}:
The non-monotonic logic programming paradigm Answer Set Programming (ASP)~\citep{gelfond_logic_2002,schaub_special_2018} has seen a rise in popularity in industry~\citep{abels_train_2021}.
A program is written as a set of rules, which is first grounded~\cite{kaminski_foundations_2023} and then solved~\cite{gebser_theory_2016}.
For details, we refer to~\citep{eiter_answer_2009}.
We translate problem (2) into ASP's syntax.
%
\begin{lstlisting}
cat(tommi). animal(X) :- cat(X).
\end{lstlisting}

\noindent \textbf{NLTK}:
The natural language toolkit~\citep{bird_natural_2009} is a Python library that enables an integration
of FOL with Prover9~\citep{mccune_prover9_2010}.
We assume familiarity with the semantics of FOL.
We show a translation of problem (2) into NLTK's syntax.
%
\begin{lstlisting}
cat(Tommi); %*all *)x. (cat(x) %*->*) animal(x))
\end{lstlisting}

\noindent \textbf{FOL}:
We assume familiarity with the syntax and semantics of FOL.
For our experiments, we implemented a \textit{parser} that translates FOL to NLTK formulas, which are then solved by Prover9.
%
Our problem (2) from above is translated as follows:
\begin{lstlisting}
cat(Tommi); %*∀*)x. (cat(x) %*→*) animal(x))
\end{lstlisting}
Observe that the formal languages of NLTK and FOL differ,
but the solvers coincide.
Further, observe that Pyke's syntax is lengthy, compared to our other formal languages,
which might impact reasoning performance.

\subsection{Datasets}
We perform experiments on three datasets.
We used one partly hand-crafted ICL-example (training data) per dataset/formal language,
which is not part of the test set.
Each test set configuration resembles the configuration of Logic-LM.
%

\noindent \textbf{ProntoQA}~\citep{saparov_language_2023}.
The ProntoQA dataset is a generated dataset.
We use the fictional character version
with a reasoning depth of 5. 
A random answer has a probability of $50\%$ for getting a correct answer (closed-world assumption - CWA), and a test set with 500 samples is used.

\noindent \textbf{ProofWriter}~\citep{tafjord_proofwriter_2021}.
ProofWriter is a generated dataset.
We chose a reasoning depth of 5.
A random answer has a probability of about $33\%$ to get a correct answer (open-world assumption - OWA).
The test set has 600 samples.

\noindent \textbf{FOLIO}~\citep{han_folio_2024}.
FOLIO is a (partly) expert-written dataset.
A random answer is correct with about $33\%$ (OWA).
%
The FOLIO test set has 204 samples.
As ASP and Pyke are logic programming paradigms, which can only handle a (proper) subset of first-order logic,
we do not use them on the FOLIO dataset.

\begin{figure*}[t]
    \includegraphics[width=15.5cm]{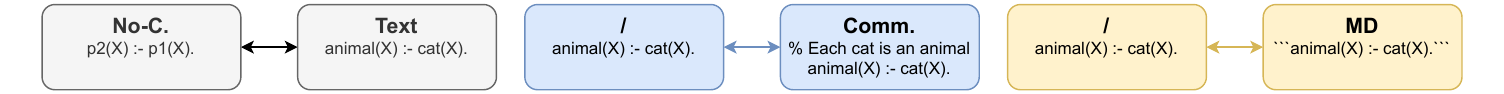}
    \vspace{-0.5em}
    \caption{
        Ablation study design about ICL-example encodings on 3 axes.
        Comparing \textit{context} with \textit{No-C.} and \textit{Text},
        \textit{comments} with \textit{/} and \textit{Comm.}, and
        \textit{markdown} with \textit{/} and \textit{MD}.
        Examples shown in ASP syntax.
    }
    \label{sec:logic-lm:fig:supplementary-information}
    \vspace{-1em}
\end{figure*}
\begin{figure*}[t]
    \begin{subfigure}{0.5\textwidth}
        \includegraphics[width=7.7cm]{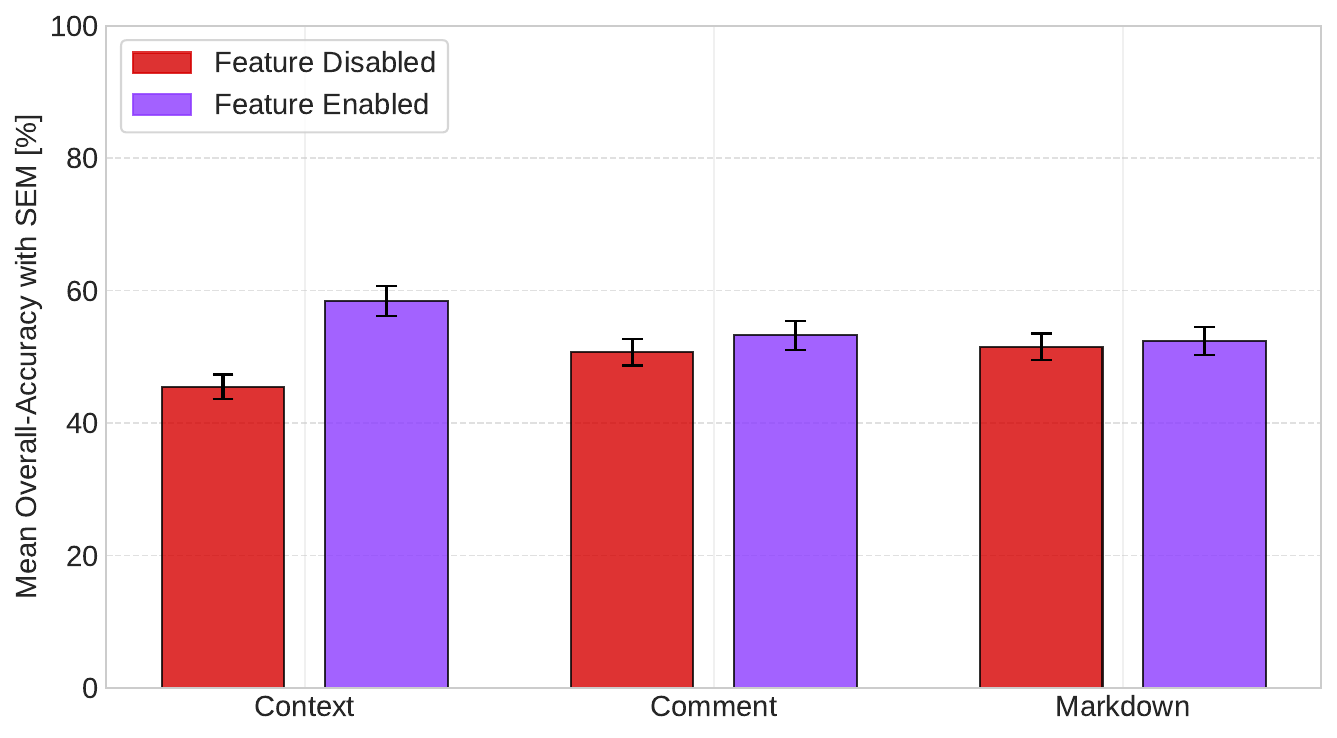}
    \end{subfigure}
    \begin{subfigure}{0.5\textwidth}
        \includegraphics[width=7.7cm]{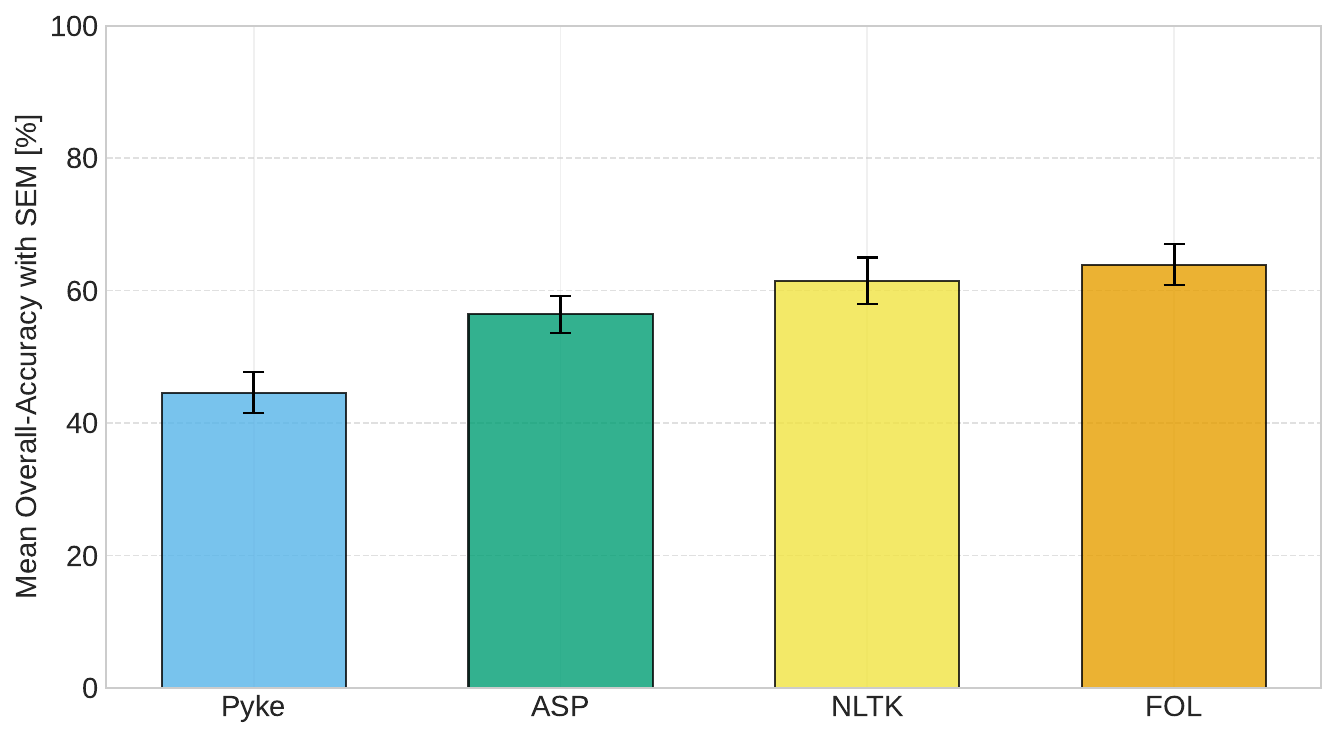}
    \end{subfigure}
    \vspace{-1.0em}
    \caption{
    We show the effects of the ablation study (left), averaged across all formal languages, LLMs, and datasets
    and the effects of the formal languages (right), averaged across all ablation study scenarios, LLMs, and the ProntoQA and ProofWriter datasets.
    Error bars show the SEM, where $n=200$ (left) and $n=80$ (right).
    }
    \label{fig:max-results-folio-ablation}
    \vspace{-0.5em}
\end{figure*}

\vspace{-0.3cm}
\subsection{Large Language Models}
\vspace{-0.1cm}

We compare the formal languages on 6 different large language models, ranging from 8 billion to 671 billion parameters.
For all experiments we set the temperature to 0, to obtain a near-deterministic behavior.
We restricted the maximum number of new tokens to be 2048
and did not perform any additional modifications to the LLMs.

%
We are primarily interested in how the intermediate language affects small models ($\approx$ 8 billion parameters).
This comes, as we view neurosymbolic AI as a potential enabler for low-resource LLM usage.
%
%
We used the following LLMs of approximately 8B parameters:
\textit{GPT-4o-mini}\footnote{\url{https://platform.openai.com/docs/models/gpt-4o-mini}},
\textit{Ministral-8B}\footnote{\url{https://mistral.ai/news/ministraux}},
\textit{Llama-8B}\footnote{\url{https://openrouter.ai/meta-llama/llama-3.1-8b-instruct}}.
and \textit{DeepSeek-8B}\footnote{\url{https://openrouter.ai/deepseek/deepseek-r1-distill-llama-8b}}.

To study the effects when using bigger models,
we additionally perform experiments on 
\textit{DeepSeek-32B}\footnote{\url{https://openrouter.ai/deepseek/deepseek-r1-distill-qwen-32b}} ($\approx$ 32 billion parameters)
and \textit{DeepSeek-V3}\footnote{\url{https://api-docs.deepseek.com/news/news1226}} ($\approx$ 671 billion parameters) models.

\vspace{-0.2cm}
\subsection{Baselines}
We show for each dataset the \textit{chance} of getting a correct answer by a random draw.
Chance is either $50\%$ for ProntoQA, as it has a CWA,
or $33\%$ for ProofWriter and FOLIO, as they have an OWA.
Additionally, we use the following four baselines.

\noindent \textbf{Standard} - 
refers to standard prompting.
The LLM is given a short instruction on the task,
the natural language-posed problem,
and a short example of how the LLM shall answer the question.

\noindent \textbf{CoT} - 
refers to CoT prompting.
The LLM is given a short instruction on the task,
the natural language-posed problem,
and an example which showcases 
how to use CoT for obtaining a correct solution.

\noindent \textbf{Logic-LM*} and \textbf{LINC*} -
refer to the formal languages of Logic-LM and LINC.
We do not use any auxiliary systems of either Logic-LM or LINC.
As formal languages,
Logic-LM uses a custom Pyke derivative for ProntoQA and ProofWriter
and a slightly adapted FOL encoding for FOLIO.
LINC uses NLTK as a formal language.
We use the ICL-instructions of Logic-LM and LINC in user-mode.
\begin{figure*}[t]
    \begin{subfigure}{0.5\textwidth}
        \includegraphics[width=7.7cm]{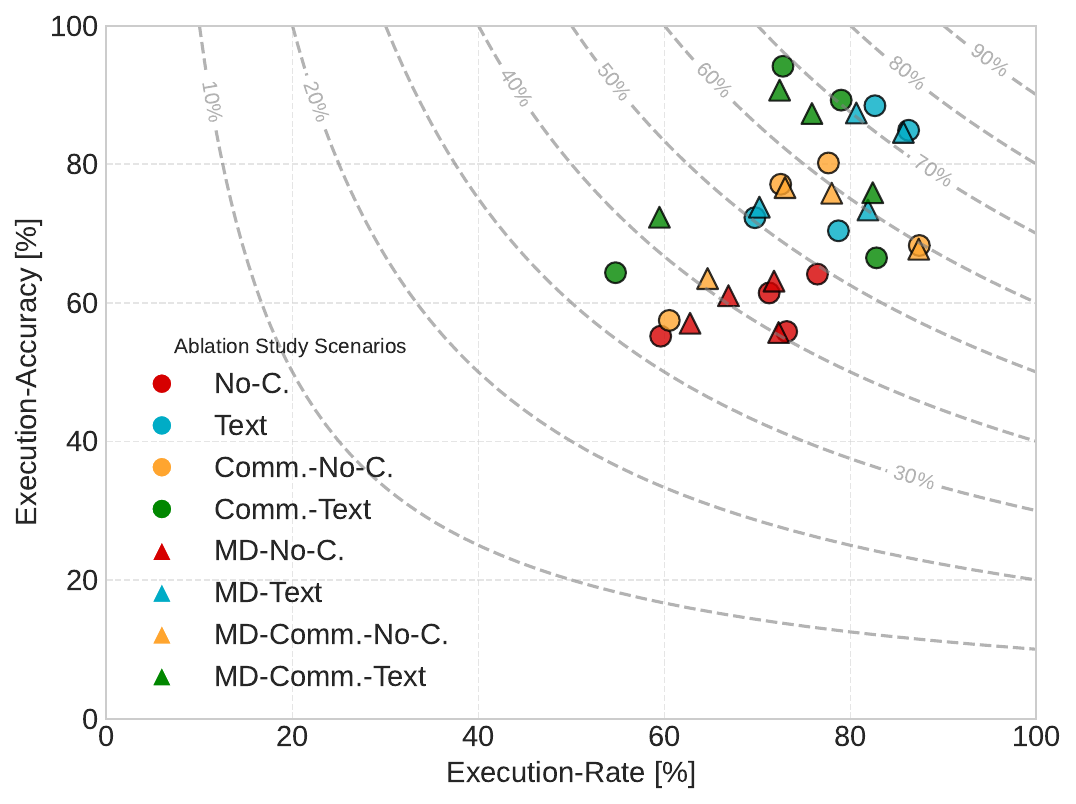}
    \end{subfigure}
    \begin{subfigure}{0.5\textwidth}
        \includegraphics[width=7.7cm]{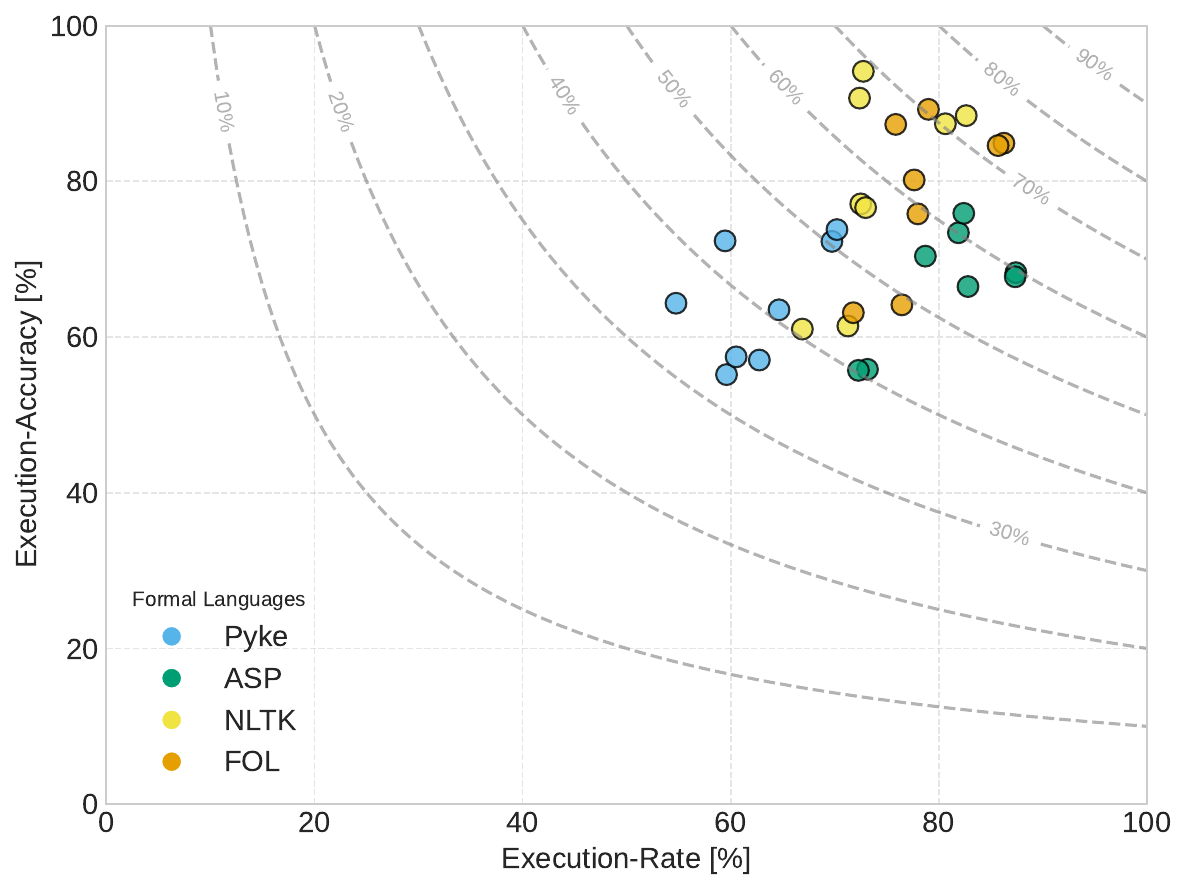}
    \end{subfigure}
    \vspace{-0.7cm}
    \caption{
    Scatter plots comparing execution-rate to execution-accuracy for the $8$ ablation study scenarios (left)
    and the formal languages (right).
    Both are averaged across ProntoQA and ProofWriter datasets and all LLMs ($n=10$).
    Contour lines show overall-accuracy in steps of $10\%$.
    }
    \label{fig:exec-rate-vs-exec-acc}
    \vspace{-0.8em}
\end{figure*}

\begin{table}[t]
    \centering
{\small
    \begin{tabular}{c|c|c||c|c|c}
        Scenario & Avg. & SEM &  Lang. & Avg. & SEM \\ \cmidrule{1-3} \cmidrule{4-6}
        No-C. & 45.49 & 1.84 & Pyke & 44.55 & 3.09  \\\cmidrule{4-6} 
        Text & \textbf{58.42} & 2.22 & ASP & 56.40 & 2.84  \\ \midrule
        / & 50.67 & 1.98 & NLTK & 61.44 & 3.54 \\\cmidrule{4-6}
        Comm. & 53.23 & 2.18 & FOL & \textbf{63.92} & 3.08 \\ \midrule
        / & 51.49 & 2.03 \\
        MD & 52.42 & 2.14 \\ \midrule
    \end{tabular}
    }
    \vspace{-0.5em}
    \caption{Scenarios per formal language ablation study (left)
    and formal languages overall results (right).
    All values in [\%], for average overall-accuracy (Avg.) and SEM, with $n=200$ (left) and $n=80$ (right).
    }
    \label{tab:ablations-study-scenarios-per-formal-language}
\end{table}

\vspace{-1.5cm}
\subsection{Ablation Study Design}
%
In addition to the choice of the formal language,
we are interested in studying how different ICL-example encodings affect the reasoning performance.
Both Logic-LM and LINC encode predicates names in a way that is close to the semantic concept.
Example: ``Tommi is a cat'' is translated as $cat(tommi)$ and not $p1(tommi)$.
Further, Logic-LM and LINC include comments for every rule, respectively formula.
These comments, written in natural language, relate the rule or formula to the natural language-posed problem.
As syntax affects the reasoning performance of LLMs,
these encoding choices made by Logic-LM and LINC have an effect on reasoning performance.

To shed light on the effects,
we study \textit{what} the effects of predicate naming (context),
the inclusion of comments (comm.), and the wrapping of problems in markdown syntax (MD) are.
These are our three axes of variation for our ablation study, resulting in \emph{8} scenarios.
We show the schematics of the ablation study design in Figure~\ref{sec:logic-lm:fig:supplementary-information}.

\noindent \textbf{Context}:
We measure the impact of predicate names on reasoning performance.
Predicates are either given a suitable name according to the problem definition (Text)
or are enumerated (No-C.).
%

\noindent \textbf{Comments}:
We study whether adding comments (Comm.) to the ICL-example, such as in Logic-LM and LINC,
actually helps, or not (/).
%

\noindent \textbf{Markdown}:
Code, or formal language, is often wrapped in markdown code syntax by LLMs.
We test whether putting the example encoding in markdown code syntax has an effect (MD), or not (/).

\subsection{Experimental Evaluation}

%
We conduct our experiments on an adapted Logic-LM implementation.
Our adaptation includes an ASP symbolic solver based on Clingo~\citep{gebser_theory_2016},
a new Pyke implementation, and an adapted NLTK/FOL solver implementation.
%
We conduct experiments for $4$ formal languages and $8$ scenarios per formal language, leading to $32$ total experiments for ProntoQA and ProofWriter.
Including the $4$ baseline experiments, we report $36$ experiments, respectively.
For FOLIO, we conduct $20$ experiments in total (Pyke and ASP cannot be measured).
This leads to a total of $92$ experiments per LLM and $552$ experiments
with $39280$ queries passed to LLMs.
We report costs of about $110\$$.

Let $\#D$ be the dataset size,
$\#\text{EXEC}$ the number of correctly parsed instances, and
$\#\text{TRUE}$ the number of correctly solved instances.
\emph{Syntactically correct} refers to a translation that adheres to the defined formal language,
whereas \emph{correctly solved} refers to a correct syntactical translation and the correct output of the solver.
%
%
The \textit{execution-rate} is the fraction of correct syntactical outputs (Exec-Rate, $\frac{\#\text{EXEC}}{\#D}$),
\textit{execution-accuracy}, is the fraction of correctly solved instances of all syntactically correct ones (Exec-Acc, $\frac{\#\text{TRUE}}{\#\text{EXEC}}$),
and \textit{overall-accuracy} 
is the fraction of correctly solved instances over the entire dataset (Overall-Acc, $\frac{\#\text{TRUE}}{\#\text{\#D}}$).
Observe: $\textit{Overall-Acc} = \textit{Exec-Acc} \cdot \textit{Exec-Rate}$.

Logic-LM and LINC employ backup procedures to increase overall-accuracy.
Backup procedures, such as CoT prompting on syntax errors, can also be integrated with our formal languages.
Baselines not using neurosymbolic reasoning (i.e., Standard and CoT) are considered to have an execution-rate of $100\%$,
while their execution-accuracy resembles their overall-accuracy,
as they are not required to adhere to a formal language.

\section{Results}
\label{section:experiments:experiments}

\begin{figure}[t]
        \includegraphics[width=7.5cm]{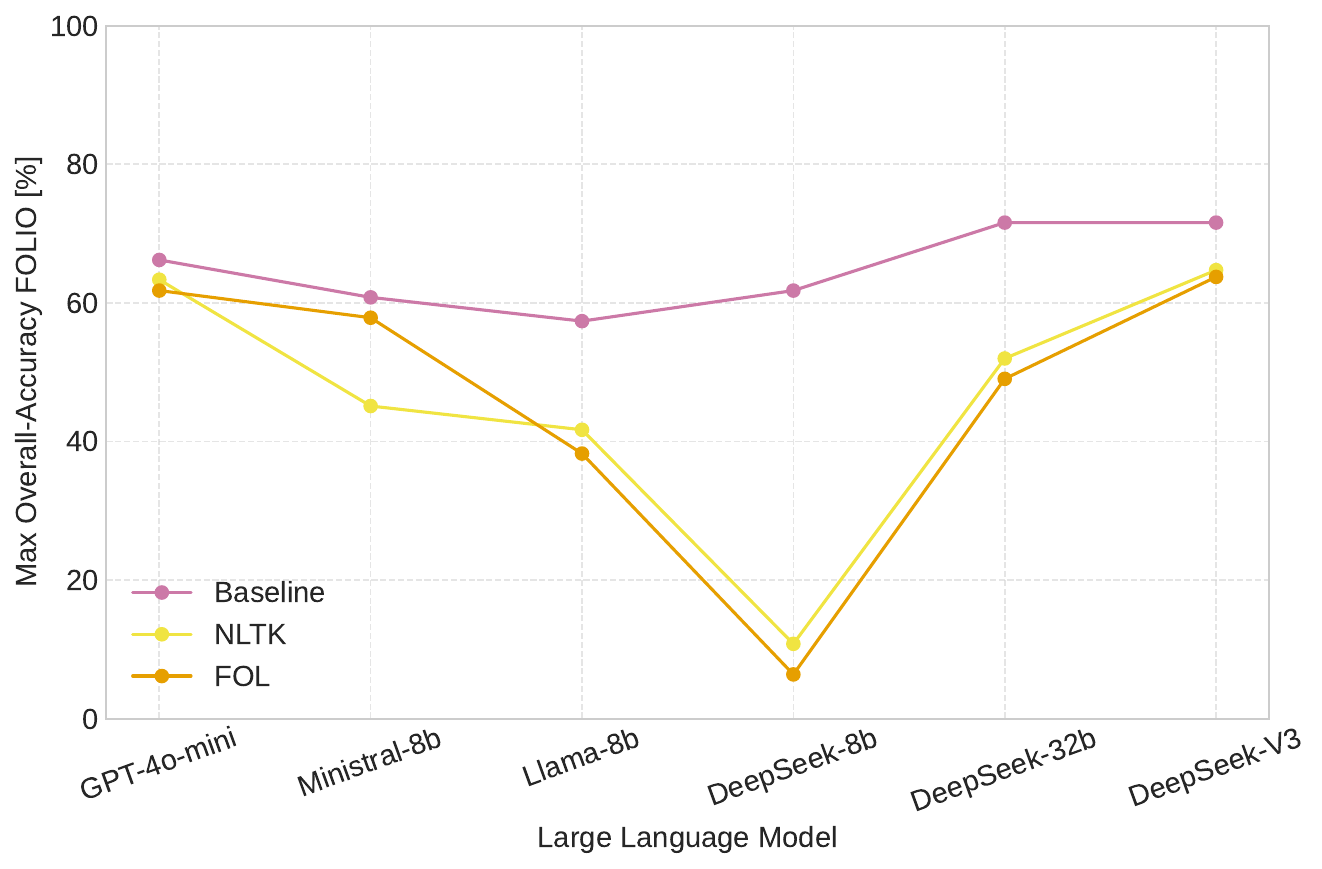}
        \caption{
        Par.-coord. plot showing max overall-accuracy for the FOLIO dataset for all LLMs.
    Maximum is computed w.r.t. all scenarios per formal language.
        }
        \label{fig:max-results-folio}
\end{figure}

We show the experimental results in Figures~\ref{fig:max-results-prontoqa-proofwriter}--\ref{fig:max-results-folio} 
and Tables~\ref{tab:ablations-study-scenarios-per-formal-language}--\ref{tbl:ministral-8b-results}.
We perform the Wilcoxon signed rank test with a significance level of $0.01$,
where we show the details in the appendix.
Here we present the main findings.
%

Figure~\ref{fig:max-results-prontoqa-proofwriter} shows the best (max) overall-accuracy each formal language achieved per LLM,
for the ProntoQA and ProofWriter datasets.
Performances vary greatly between LLMs and formal languages.
Generally, the best results of FOL beat the best ones of ASP and Pyke on both datasets for all LLMs.
%
For DeepSeek-32B and DeepSeek-V3, the best neurosymbolic approaches were able to beat (or be equal to) the baseline in three out of four cases.
For FOLIO (Figure~\ref{fig:max-results-folio}), the neurosymbolic approaches show promising results,
as they approach the performance of the baselines within $10\%$ for 3 LLMs.

In Figure~\ref{fig:max-results-folio-ablation} and Table~\ref{tab:ablations-study-scenarios-per-formal-language}, we show averaged results with the standard error of the mean (SEM) for the ablation study scenarios per formal language and
the effects of the different formal languages.
We average an ablation study scenario over all LLMs, datasets, and formal languages.
For averaging the formal languages, we compute the average across all LLMs, ablation study scenarios,
and the datasets ProntoQA and ProofWriter.

While the use of context increases overall-accuracy, the results for using comments or wrapping ICL-examples in markdown code are inconclusive.
The best results were achieved using text and at least one of comments or markdown code.
In general, FOL achieves better results than ASP or Pyke.
Further, NLTK and ASP have a higher overall-accuracy than Pyke.
The results between FOL and NLTK, and NLTK and ASP, are inconclusive.
%
For the problems in the datasets, we do not encounter problems when solving in terms of \emph{intractability} - a combinatorial explosion in the solver.
Therefore, we are not required to use special strategies for tackling intractability,
such as symmetry breaking~\cite{fahle_symmetry_2001} or tackling the ASP bottleneck~\cite{beiser_bypassing_2024}.

\begin{table*}[t]
    \centering
    \resizebox{15.1cm}{!}{
    \centering
    \begin{tabular}{ll|ccc|ccc|ccc}
    \toprule
        \multicolumn{2}{c}{Method} & \multicolumn{3}{c}{ProntoQA} & \multicolumn{3}{c}{ProofWriter} & \multicolumn{3}{c}{FOLIO} \\

        \cmidrule(lr){3-5} \cmidrule(lr){6-8} \cmidrule(lr){9-11}   &  & Overall-Acc & Exec-Rate & Exec-Acc & Overall-Acc & Exec-Rate & Exec-Acc & Overall-Acc & Exec-Rate & Exec-Acc\\
    \midrule
Chance &  & 50.00 & / & / & 33.33 & / & / & 33.33 & / & / \\ \midrule
\multirow{4}{*}{Baseline} & Standard & 48.80 & 100.00 & 48.80 & 47.33 & 100.00 & 47.33 & 54.90 & 100.00 & 54.90 \\
 & CoT & \textit{86.60} & 100.00 & 86.60 & \textit{51.67} & 100.00 & 51.67 & \textbf{60.78} & 100.00 & 60.78 \\
 & Logic-LM* & 2.20 & 3.20 & 68.75 & 0.00 & 0.00 & 0.00 & 0.00 & 0.00 & 0.00 \\
 & LINC* & 6.20 & 6.80 & 91.18 & 5.00 & 13.67 & 36.59 & 1.96 & 8.82 & 22.22 \\ \midrule
\multirow{8}{*}{Pyke} & No-C. & 44.00 & 89.20 & 49.33 & 18.17 & 45.83 & 39.64 & / & / & / \\
 & Text & 66.20 & 97.20 & 68.11 & 50.17 & 84.33 & 59.49 & / & / & / \\
 & Comm.-No-C. & 45.20 & 72.00 & 62.78 & 21.50 & 46.67 & 46.07 & / & / & / \\
 & Comm.-Text & 38.80 & 52.40 & 74.05 & 15.17 & 23.17 & 65.47 & / & / & / \\
 & MD-No-C. & 47.60 & 99.20 & 47.98 & 11.33 & 26.50 & 42.77 & / & / & / \\
 & MD-Text & \textit{69.60} & 97.80 & 71.17 & \textit{52.00} & 86.17 & 60.35 & / & / & / \\
 & MD-Comm.-No-C. & 61.20 & 99.00 & 61.82 & 21.67 & 41.83 & 51.79 & / & / & / \\
 & MD-Comm.-Text & 64.60 & 78.80 & 81.98 & 11.67 & 18.33 & 63.64 & / & / & / \\ \midrule
\multirow{8}{*}{ASP} & No-C. & 42.40 & 86.00 & 49.30 & 13.67 & 29.00 & 47.13 & / & / & / \\
 & Text & 65.60 & 79.40 & 82.62 & 41.83 & 73.67 & 56.79 & / & / & / \\
 & Comm.-No-C. & 61.40 & 98.60 & 62.27 & 65.50 & 94.00 & 69.68 & / & / & / \\
 & Comm.-Text & 88.20 & 89.20 & 98.88 & 48.83 & 78.83 & 61.95 & / & / & / \\
 & MD-No-C. & 30.60 & 62.40 & 49.04 & 23.83 & 49.00 & 48.64 & / & / & / \\
 & MD-Text & 86.60 & 95.60 & 90.59 & 38.33 & 70.67 & 54.25 & / & / & / \\
 & MD-Comm.-No-C. & 62.20 & 99.40 & 62.58 & \textit{66.67} & 93.17 & 71.56 & / & / & / \\
 & MD-Comm.-Text & \textit{93.20} & 93.80 & 99.36 & 50.17 & 80.67 & 62.19 & / & / & / \\ \midrule
 \multirow{8}{*}{NLTK} & No-C. & 40.40 & 87.60 & 46.12 & 32.50 & 65.83 & 49.37 & 4.90 & 12.25 & 40.00 \\
 & Text & 97.20 & 99.80 & 97.39 & 87.00 & 97.17 & 89.54 & 26.96 & 69.12 & 39.01 \\
 & Comm.-No-C. & 82.20 & 100.00 & 82.20 & 72.50 & 83.50 & 86.83 & 0.49 & 1.47 & 33.33 \\
 & Comm.-Text & \textit{99.60} & 100.00 & 99.60 & 93.00 & 97.17 & 95.71 & \textit{45.10} & 76.96 & 58.60 \\
 & MD-No-C. & 29.20 & 58.00 & 50.34 & 35.50 & 74.67 & 47.54 & 6.37 & 17.16 & 37.14 \\
 & MD-Text & 93.00 & 100.00 & 93.00 & 89.17 & 97.00 & 91.92 & 33.82 & 76.96 & 43.95 \\
 & MD-Comm.-No-C. & 82.80 & 100.00 & 82.80 & 75.50 & 85.67 & 88.13 & 0.98 & 2.45 & 40.00 \\
 & MD-Comm.-Text & 99.40 & 100.00 & 99.40 & \textit{94.17} & 97.83 & 96.25 & 44.61 & 72.55 & 61.49 \\ \midrule
\multirow{8}{*}{FOL} & No-C. & 37.80 & 81.60 & 46.32 & 36.33 & 69.50 & 52.28 & 35.78 & 67.16 & 53.28 \\
 & Text & 92.60 & 100.00 & 92.60 & 87.83 & 98.17 & 89.47 & 45.10 & 74.51 & 60.53 \\
 & Comm.-No-C. & 83.40 & 100.00 & 83.40 & 86.67 & 95.83 & 90.43 & 55.88 & 87.25 & 64.04 \\
 & Comm.-Text & \textbf{100.00} & 100.00 & 100.00 & \textbf{94.67} & 98.67 & 95.95 & \textit{57.84} & 79.41 & 72.84 \\
 & MD-No-C. & 26.60 & 57.20 & 46.50 & 35.17 & 66.17 & 53.15 & 33.82 & 70.59 & 47.92 \\
 & MD-Text & 86.40 & 100.00 & 86.40 & 86.67 & 97.67 & 88.74 & 49.02 & 79.90 & 61.35 \\
 & MD-Comm.-No-C. & 82.60 & 100.00 & 82.60 & 87.50 & 96.17 & 90.99 & 57.35 & 88.24 & 65.00 \\
 & MD-Comm.-Text & 99.60 & 100.00 & 99.60 & 91.17 & 95.50 & 95.46 & \textit{57.84} & 78.92 & 73.29 \\
    \bottomrule
    \end{tabular}
        }
        \caption{
        Detailed results for the Ministral-8B model, depicting overall-accuracy, execution-rate, and execution-accuracy for the ProntoQA, ProofWriter, and FOLIO datasets.
        All values shown in percent [\%].
        }
        \label{tbl:ministral-8b-results}
\end{table*}



In Figure~\ref{fig:exec-rate-vs-exec-acc}, we show scatter plots of the execution-rate (x-axis) vs. execution-accuracy (y-axis),
for the ablation study scenarios and the formal languages.
Both plots show the same data, however, with different legends.
Each dot represents a formal language with a certain ablation study scenario, averaged across all LLMs and ProntoQA, and ProofWriter datasets.
The overall-accuracy is obtained by multiplying a point's x-position with its respective y-position.

No context leads to both lower execution-accuracy and lower execution-rate,
as
%
out of the $12$ points below $50\%$ overall-accuracy $10$ have no-context.
The best results are achieved with text and text with markdown.
%
%
%
Pyke performs approximately equally well on execution-rate and execution-accuracy,
while having all dots below the $60\%$ overall-accuracy mark,
where most (6 out of 8) are below $50\%$.
ASP's execution-rate tends to stay relatively high,
by ranging from $72.27\%$ to $87.41\%$,
%
and its overall-accuracy does not fall below $40\%$, while it goes beyond $60\%$.
NLTK's execution-accuracy is relatively high,
by staying above $60\%$. 
Its overall-accuracy does not fall below $40\%$ 
and reaches over $70\%$.
%
FOL's behavior is similar to NLTK's, however, with a higher execution-rate,
resulting in 
FOL's overall-accuracy being marginally higher than NLTK's.

\subsection{Qualitative Error Analysis}
\label{section:qualitative-error-analysis}
In this section, 
we discuss common errors across formal languages to get a better understanding of the diffferences in our statistical results.

Across all formal languages, we experience cases where the LLMs seemingly got trapped in an endless output token generation,
which is only stopped by setting a hyperparameter that caps the maximum number of output tokens (2048 in our case).
For \emph{Pyke} in particular, we notice that LLMs format the output incorrectly, by missing line breaks.
When translating to \emph{ASP}, LLMs struggle to distinguish the two notions of negation:
\emph{strong}, written as $-$, and \emph{default}, written as $not$.
This results in program statements such as \textit{-not p1(wren)}, which are syntactically incorrect.
The syntactic errors between \emph{NLTK} and \emph{FOL} are similar.
Examples include incorrectly setting parentheses or using a predicate with multiple arities.
For example, using $p14(X)$ and $p14(X,Y)$. 
The Logic-LM* and LINC* neurosymbolic baselines were particularly prone to small syntax errors,
like wrapping lines or predicates in markdown bold-faced letters, or enumerating lines.
Take, for example, the intended output of \textit{Cold(\$x,bool)} and the produced output \textit{1. **Cold(\$x, bool)**}.
We consider such translations as syntactically false, which explains the bad performance of Logic-LM* and LINC* in Table~\ref{tbl:ministral-8b-results}.

%
%
%
\vspace{-0.1cm}
\section{Related Work}
\label{sec:related-work}
\vspace{-0.1cm}

%
%
%
%
Improving LLM's reasoning capability was approached by different angles.
CoT prompting,
part of the emergent ICL or \textit{few-shot-learning} capability~\citep{shanahan_talking_2024},
improves LLMs performance on reasoning tasks~\citep{wei_chain--thought_2022}.
However, CoT is non-faithful~\citep{lyu_faithful_2023} and
LLMs ``remain limited in their capabilities to performing probabilistic retrieval''~\citep{panas_can_2024}.
Related results show that LLMs do not
acquire systematic problem solving skills~\citep{dziri_faith_2023}.
Fine-tuning or pre-training improves numerical capabilities~\citep{geva_injecting_2020},
syntax recognition of ASP with LLASP~\citep{coppolillo_llasp_2024},
or proof verification~\citep{geva_injecting_2020}.
However, it remains unclear whether fine-tuning enables LLMs to reason precisely~\citep{panas_can_2024}.

Neurosymbolic AI~\citep{garcez_neurosymbolic_2023} combines  approaches offer an alternative to the pure sub-symbolic approaches,
ranging from differentiable logic~\citep{badreddine_logic_2022}
over visual question answering~\citep{eiter_logic-based_2023},
to LLMs~\citep{pan_logic-lm_2023}.
%
%
For knowledge-based systems, previous research focused on the proficiency of LLMs on formal languages~\citep{liu_how_2024}.
While autoformalization~\citep{wu_autoformalization_2022} is concerned with the translation of natural language into a particular formal language, the intermediate language challenge is about choosing a suitable formal language.
For logical reasoning tasks, 
\textit{Logic-LM}~\citep{pan_logic-lm_2023} is a neurosymbolic method that
combines LLMs with symbolic solvers.
Logic-LM uses \textit{Pyke}~\citep{frederiksen_applying_2008} for logic programming,
\textit{Z3}~\citep{de_moura_z3_2008} for SAT problems,
\textit{Prover9}~\citep{mccune_prover9_2010} for FOL problems, and
the \textit{Python-constraint}~\citep{niemeyer_python-constraint_2024} library for constraint programming.
Logic-LM's formal languages deviate from the corresponding formal languages of the solvers, thereby requiring parsers.
The \textit{LINC}~\citep{olausson_linc_2023} system uses NLTK as an intermediate language
and uses multiple ICL-examples and parallel prompts.
Also, integration of the error messages of the solvers~\citep{kirtania_logic-lm_2024} into the translation process or
the usage of different solvers~\citep{lam_closer_2024} is discussed.
However, to the best of our knowledge, no neurosymbolic LLM approach thus far studied the effects of different formal languages on logical reasoning performance.

\vspace{-0.1cm}
\section{Conclusion}
\label{sec:discussion}
\vspace{-0.1cm}
Logical reasoning tasks pose a problem to LLMs,
as they remain limited in their ability to perform probabilistic retrieval~\cite{panas_can_2024}.
Neurosymbolic approaches help,
by constraining the probabilistic nature to the translation step of a 
natural language-posed problem into a formal language~\citep{pan_logic-lm_2023,olausson_linc_2023}.
Therefore, the reasoning step itself is not affected by the probabilistic nature of LLMs.

In this paper, we discuss the effect of the chosen formal language on a model's reasoning performance.
We introduce the \textit{intermediate language challenge},
which refers to the problem of choosing a suitable formal language for neurosymbolic reasoning.
In our experiments, we compare Pyke, ASP, NLTK, and FOL as formal languages,
and 8 ablation study scenarios on ICL-example encodings, using 6 different LLMs and three datasets (i.e., ProntoQA, ProofWriter, and FOLIO).
Results show that, on average, FOL performs better than logic programming approaches.
Further, FOL slightly outperforms NLTK on average,
whereas ASP clearly outperforms Pyke.
The investigated ablation study scenarios show that, on average, more content information increases overall-accuracy for neurosymbolic reasoning.
%

For future work, we want to investigate the impact of fine-tuning LLMs for neurosymbolic reasoning and study the behavior of other formal languages or prompting techniques.
The latter extends to the application on additional datasets, possibly including non-classical non-monotonic logic.

\clearpage
\section*{Limitations}
\label{sec:limitations}
To the best of our knowledge, no study thus far has compared the impact of the chosen formal language on neurosymbolic LLM reasoning performance.

\textbf{Different LLMs}.
We considered an LLM from a user perspective,
thereby being unaffected by additional training data.
In this context, we studied the behavior of formal languages primarily on small ($\approx$ 8 billion parameters) models,
while we also performed analysis also on bigger models ($\approx$ 32 billion and $\approx$ 671 billion).
Therefore, on fine-tuned, bigger, or future models, the performance might change significantly.

\textbf{Different Formal Languages.}
We considered 4 formal languages, which we chose either due to their recent usage in related neurosymbolic reasoning tasks (Pyke, NLTK, FOL),
or due to their popularity in industry or science (ASP).
However, we acknowledge that encodings in other formal languages, such as description logics, might change the results significantly.
%
The formal languages presented thus far have in common that they are declarative.
Conversely, using procedural formal languages, such as the widely used programming language of Python,
might be interesting as a comparison.
However, due to their procedural nature, it is unlikely that their usage yields substantial performance improvements on logical reasoning problems w.r.t. overall-accuracy.

\textbf{Other contributing factors affecting performance}.
The formal language is not the only factor affecting performance w.r.t. overall-accuracy.
Undoubtedly, model architecture and training data have an effect.
Although highly interesting, we consider a detailed evaluation of the model architecture and training data as outside of the scope of this study.

Further, we observed that alterations of the ICL-instruction, not only the ICL-example, has effects on overall-accuracy.
Therefore, our approach was to achieve maximal comparability by keeping ICL instructions largely the same between formal languages.
Still, we cannot rule-out the possibility that better performance is achieved when other ICL-instructions are used.
On a related note, some methods, such as LINC, instruct the LLM in \textit{system-mode} to adhere to another personality - e.g., not being a chatbot, 
but a translation engine.
We opted for a user-mode, as by doing that, a fair comparison between neurosymbolic and CoT/standard prompting baselines is ensured.
Using LLMs with a personality defined in system mode might change the results.

%
%
Retrieval-Augmented-Generation (RAG) enables the LLM to look up facts, such as syntactic definitions.
Although using RAG in a reasoning setting is related, we view it as out of scope of this study.

\textbf{Automatic Neurosymbolic Reasoning}.
Inherent to the neurosymbolic approach is the inclusion of a separate symbolic
reasoning system.
However, in an ideal integration of a symbolic solver into an LLM,
the symbolic solver's details are hidden from the end-user.
Nonetheless, the symbiosis of the LLM with the symbolic solver
into one coherent system that
\textit{automatically detects when the usage of the symbolic solver is beneficial},
might pose a major challenge.

\textbf{Selection of Logical Reasoning Tasks}.
We considered the three logical reasoning datasets ProntoQA, ProofWriter, and FOLIO.
However, the style of how questions are posed and answered is semi-formal.
Further, only certain logical reasoning aspects are considered in the datasets,
where classical logic (such as first-order logic) is the intended logical reasoning concept.
Other tasks, which would require non-classical logics such as in non-monotonic reasoning,
potentially change the reasoning performance of the systems.
As for example, ASP is a non-monotonic reasoning framework, it is expected that it performs better on non-monotonic reasoning tasks.
Further, ProntoQA and ProofWriter prominently require the usage of the modus ponens in their reasoning tasks,
which only captures a small subset of all logical reasoning tasks.

Lastly, when moving to bigger, more complex problems in the reasoning datasets,
%
additional challenges will occur that (might) prevent the usage of symbolic solvers.
Evidently, efficient neurosymbolic LLM reasoning must take this into account;
therefore, LLMs are not only required to translate the problem correctly w.r.t. syntax and semantic,
but also in a way that facilitates efficient, w.r.t time, solving by symbolic solvers.

\textbf{Additional Baselines}.
Additional baselines such as tree of thoughts~\citep{yao_tree_2023}
have the potential to improve upon traditional CoT prompting, 
thereby elevating the baseline scores.



\bibliography{submission}

\clearpage

\appendix
\section{Appendix}

In Section~\ref{sec:appendix:exp-results-details}, we show additional details of our quantitative analysis.
In Section~\ref{sec:app:qualitative-details} we show the details of the qualitative error analysis, including occurrences and examples.
In Sections~\ref{sec:app:standard-prompt}--\ref{sec:app:fol-prompt} we show example prompts.
Finally, in Section~\ref{sec:app:licences-scientific-artifacts} we discuss the licenses of the used scientific artifacts,
and in Section~\ref{sec:app:usage-of-AI-assistants} we close with a brief discussion on the usage of AI assistants.

We show in the Tables~\ref{tbl:asp-table-our-results}--\ref{tbl:deepseek-v3} additional experimental details
and in Figures~\ref{fig:ablation-study-histograms} and~\ref{fig:formal-lang-histograms} distributions relating to the averages 
of Figure~\ref{fig:max-results-folio-ablation}.
Further, we show sample prompts on the ProntoQA dataset.
All depicted prompts were prompted in user-mode, without additional information to ensure comparability between the approaches.
Besides the four baselines, we depict the 8 scenarios s.t. two scenarios are shown per formal language 
s.t. for each formal language one example contains \textit{No-C.}, while the other \textit{Text},
and additionally, at least one example contains \textit{comments}.

\subsection{Quantitative Analysis (Details)}
\label{sec:appendix:exp-results-details}

\begin{figure*}
    \centering
    \begin{subfigure}{0.49\textwidth}
        \includegraphics[width=7cm]{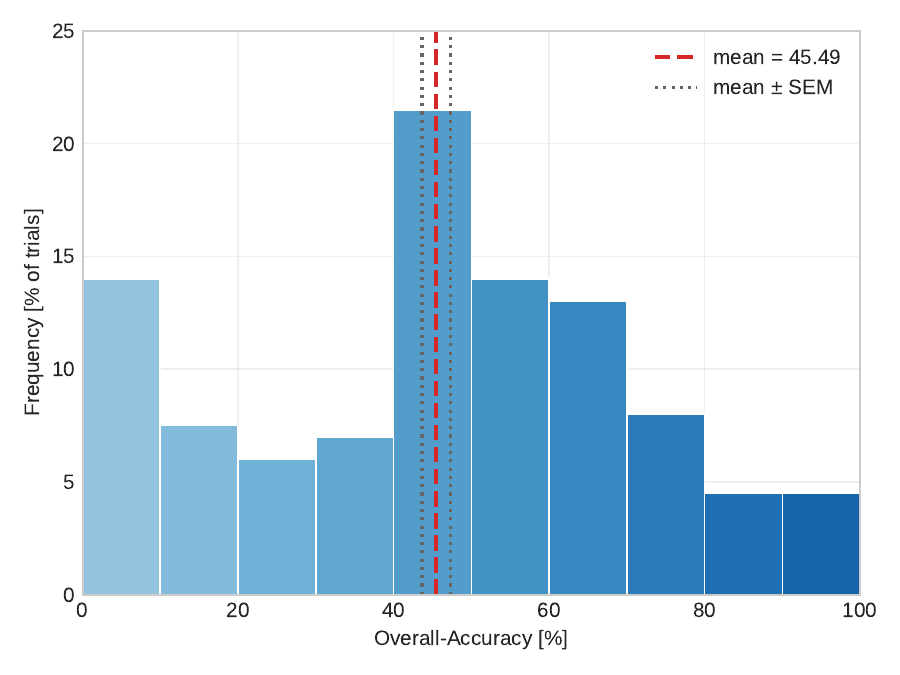}
        \caption{Histogram: No-C.}
    \end{subfigure}
    \begin{subfigure}{0.49\textwidth}
        \includegraphics[width=7cm]{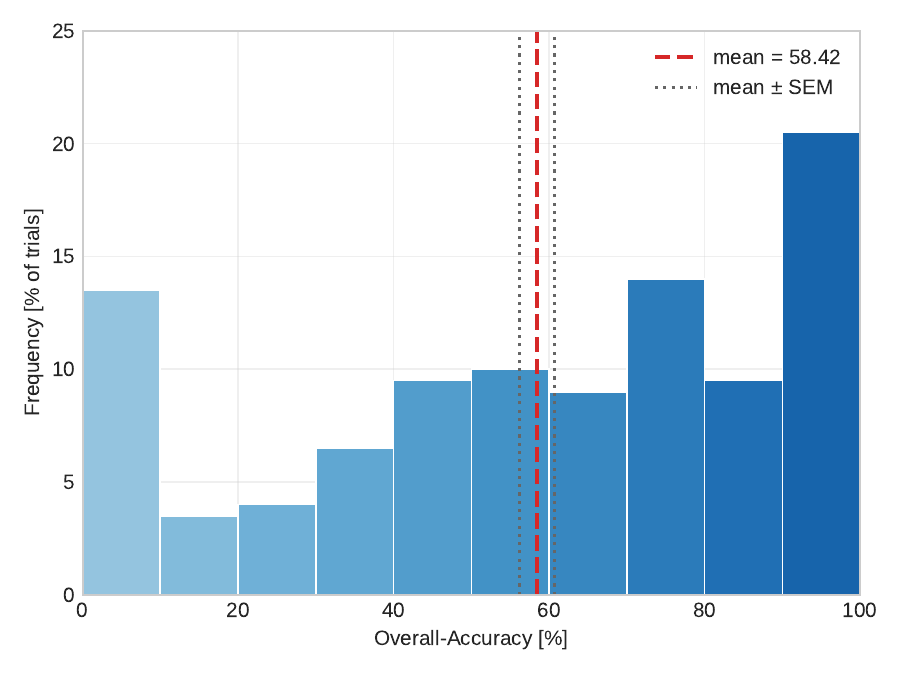}
        \caption{Histogram: Context}
    \end{subfigure}

    \centering
    \begin{subfigure}{0.49\textwidth}
        \includegraphics[width=7cm]{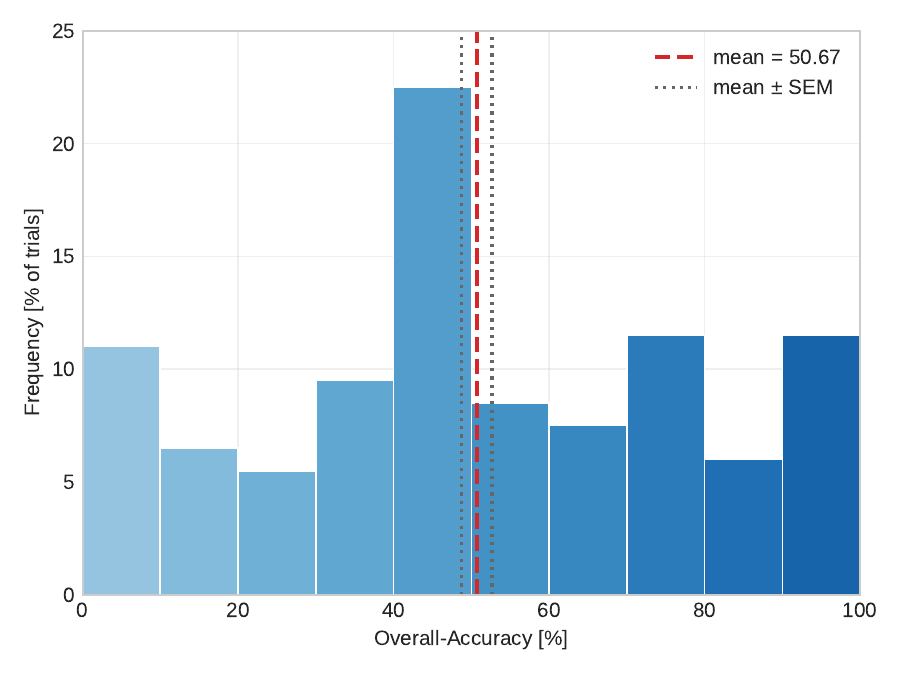}
        \caption{Histogram: No-Comment}
    \end{subfigure}
    \begin{subfigure}{0.49\textwidth}
        \includegraphics[width=7cm]{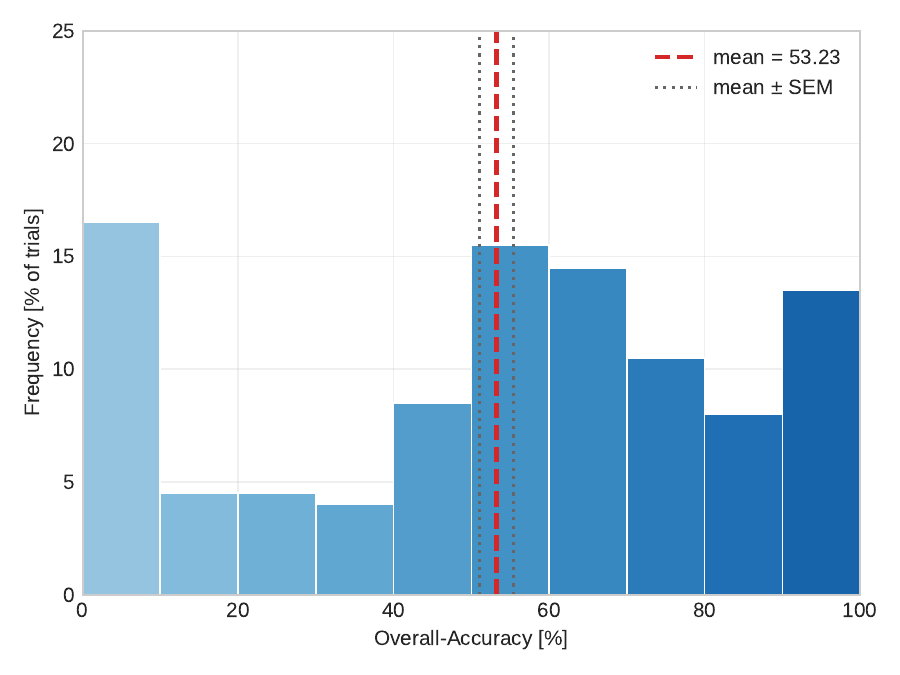}
        \caption{Histogram: Comment}
    \end{subfigure}
    
    \centering
    \begin{subfigure}{0.49\textwidth}
        \includegraphics[width=7cm]{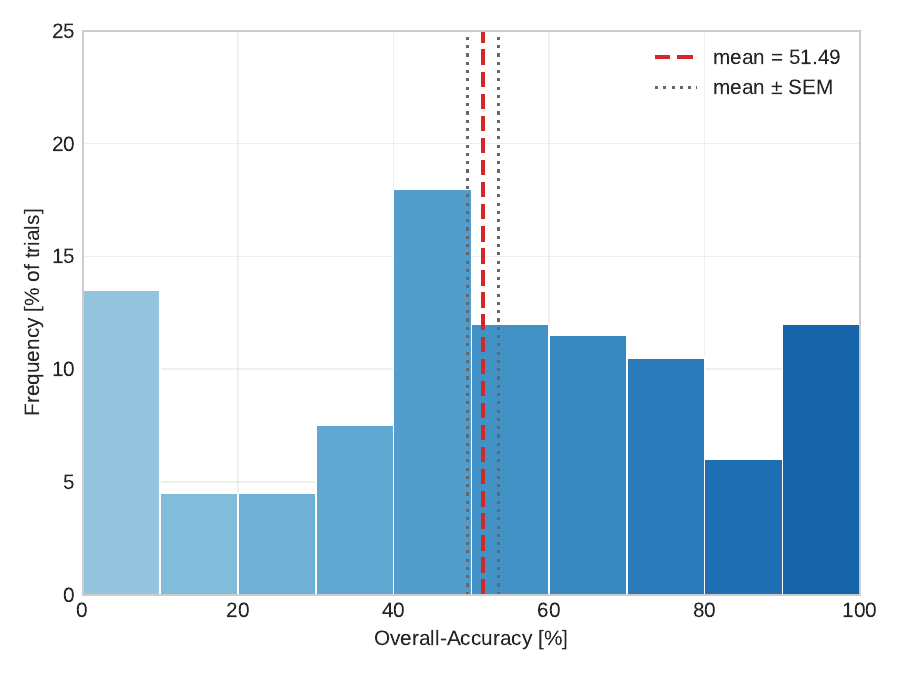}
        \caption{Histogram: No-Markdown}
    \end{subfigure}
    \begin{subfigure}{0.49\textwidth}
        \includegraphics[width=7cm]{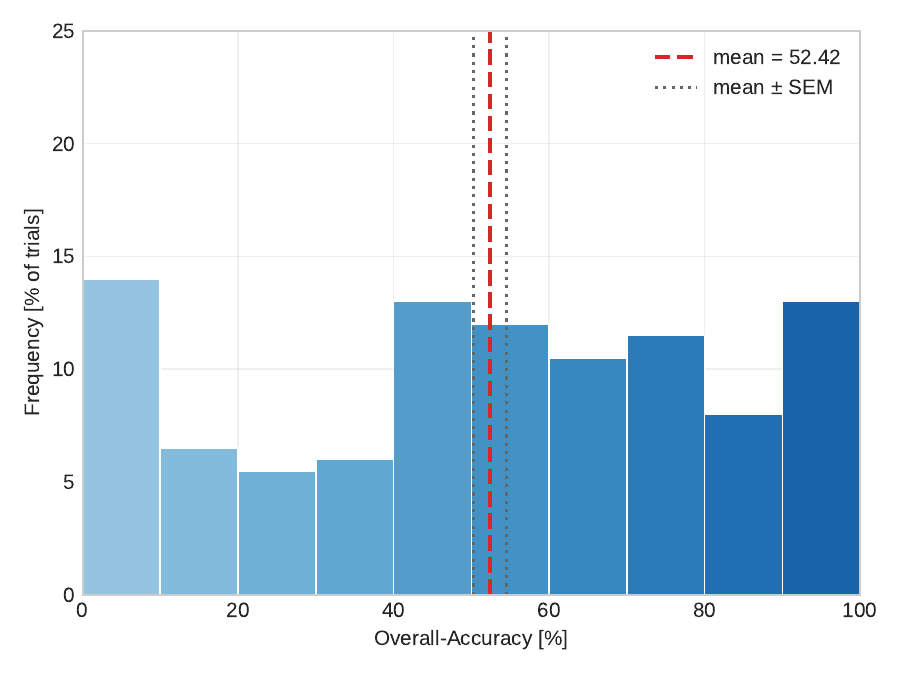}
        \caption{Histogram: Markdown}
    \end{subfigure}
    \caption{
    Histograms of the ablation study scenarios, underlying the means and standard error of the means (SEM) of Figure~\ref{fig:max-results-folio-ablation} (left).
    For all histograms, moving from the disabled feature to the enabled feature results in a slight shift of the histogram frequencies further to the right.
    }
    \label{fig:ablation-study-histograms}
\end{figure*}

\begin{figure*}
    \centering
    \begin{subfigure}{0.49\textwidth}
        \includegraphics[width=7cm]{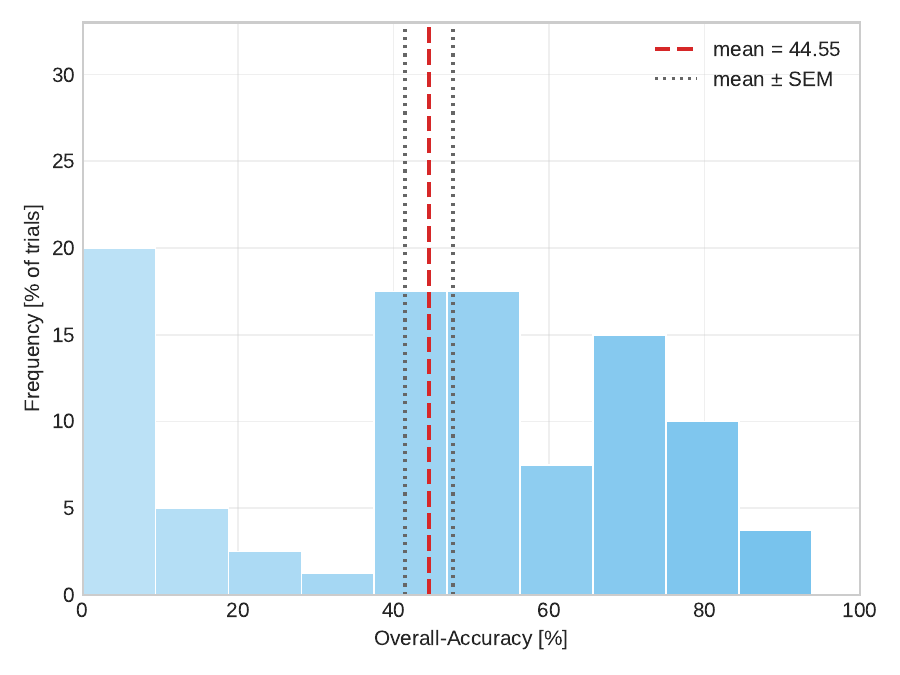}
        \caption{Histogram: Pyke}
    \end{subfigure}
    \begin{subfigure}{0.49\textwidth}
        \includegraphics[width=7cm]{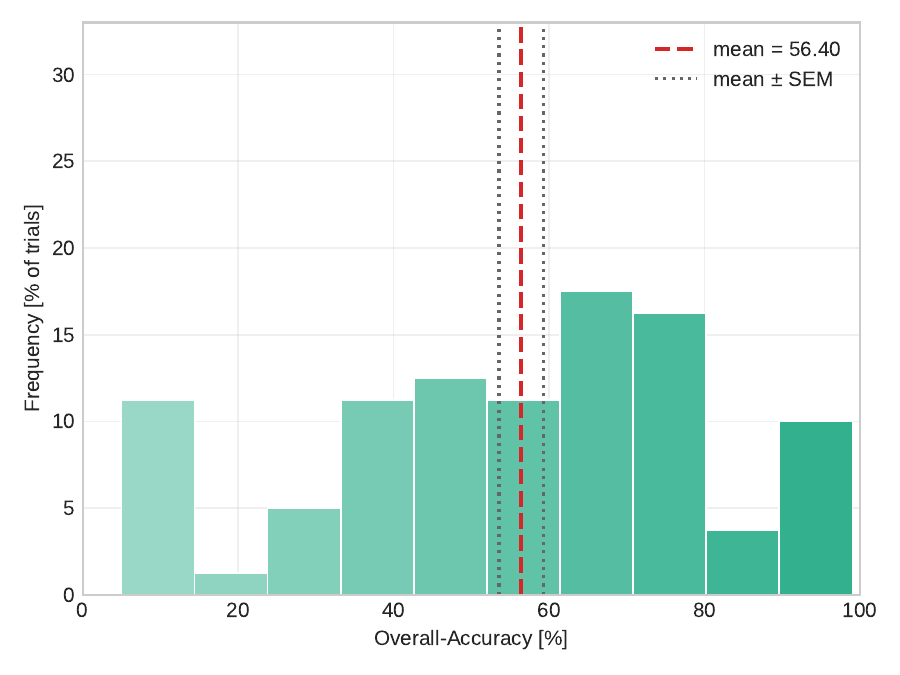}
        \caption{Histogram: ASP}
    \end{subfigure}

    \centering
    \begin{subfigure}{0.49\textwidth}
        \includegraphics[width=7cm]{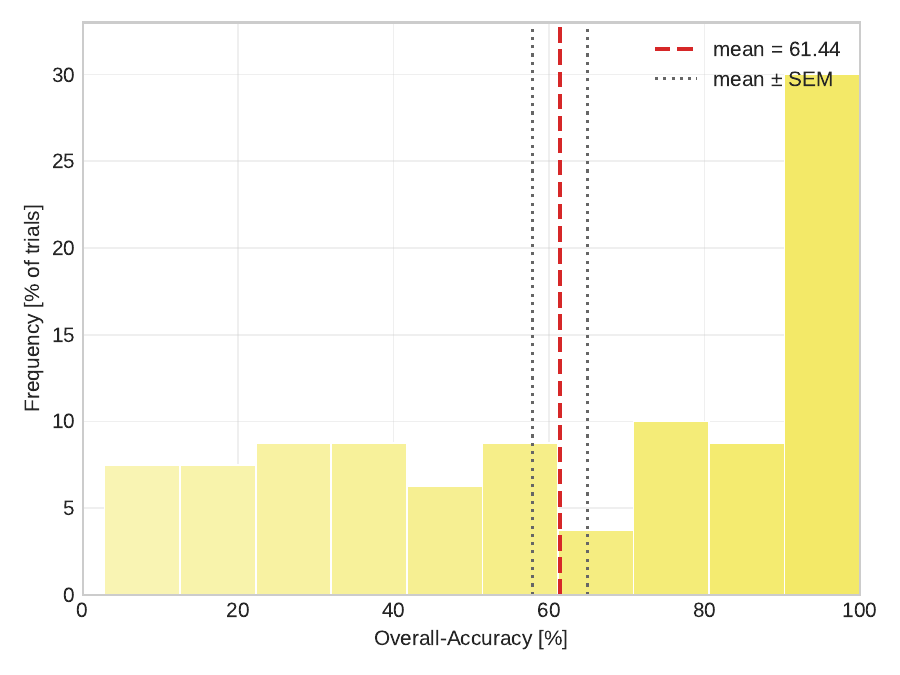}
        \caption{Histogram: NLTK}
    \end{subfigure}
    \begin{subfigure}{0.49\textwidth}
        \includegraphics[width=7cm]{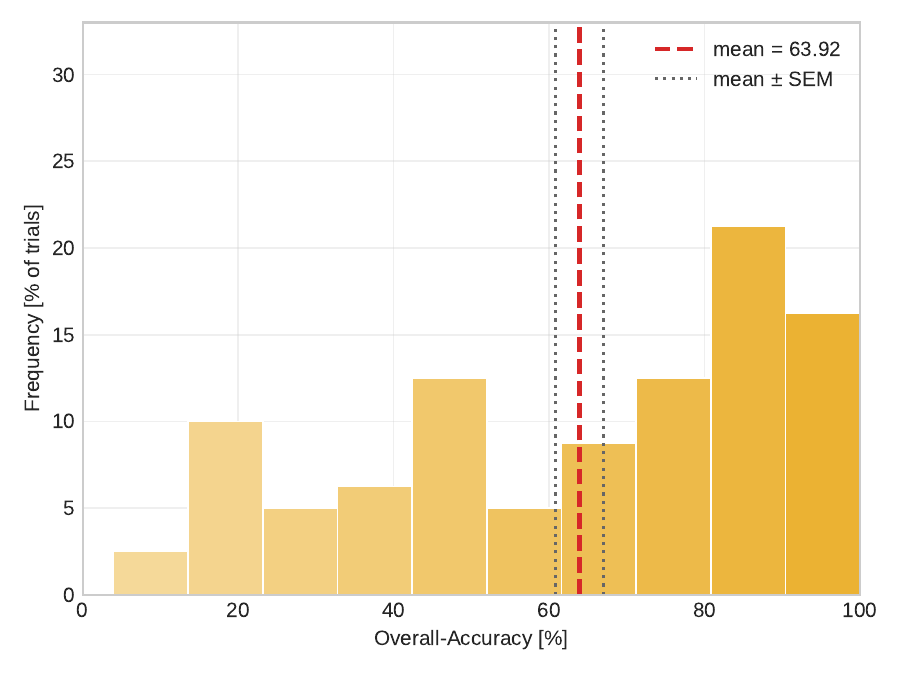}
        \caption{Histogram: FOL}
    \end{subfigure}
    \caption{
    Histograms of the formal languages underlying the means and standard error of the means (SEM) of Figure~\ref{fig:max-results-folio-ablation} (right).
    Pyke's mode is between 0 and 10\%, while a second, lower, mode is located between roughly 40 and 70\%.
    ASP's mode is located between 60 and 70\% and generally shifted to the right compared to Pyke's.
    NLTK's mode is between 90 and 100\%, however all other values are approximately evenly distributed.
    On the other hand, FOL's mode is between 80 and 90\%, while most values are shifted to the right compared to NLTK's.
    }
    \label{fig:formal-lang-histograms}
\end{figure*}

We show the results of the Wilcoxon signed rank test.
Consider the data used for Figure~\ref{fig:max-results-folio-ablation} and Table~\ref{tab:ablations-study-scenarios-per-formal-language},
where we show its distributions in Figures~\ref{fig:ablation-study-histograms} and~\ref{fig:formal-lang-histograms}.
The data is not normally distributed and it is independent per group (e.g., context, ASP, ..),
however, it is paired between ablation study groups (e.g., context and no-context)
and formal languages (Pyke, ASP, NLTK, FOL).
Therefore, we use non-parametric tests.
We choose a significance value of $\alpha = 0.01$ for few false positives.
We use the Python \textit{Scipy} package for the tests:
The function 
\texttt{wilcoxon} with parameters \texttt{alternative=``greater''} and \texttt{zero\_methods=``zsplit''}
and
the function
\texttt{friedmannchisquare} with default parameters.

\noindent
\textbf{Ablation study}.
For our ablation study, we perform the Wilcoxon signed rank test on the data tuples (feature enabled, feature disabled): (Context, No-Context), (Comment, No-Comment), (Markdown, No-Markdown).
Recall $n=200$ for each ablation study $l \in \{\text{Context},\text{No-Context},\text{Comment},\text{No-Comment},$ $\text{Markdown},\text{No-Markdown}\}$.
We perform a 1-sided test with $H_0$ being \emph{there is no difference}
and $H_1$ being \emph{feature enabled} $>$ \emph{feature disabled}.
We report the following p-values in Table~\ref{tbl:wilcoxon-ablation}.

\begin{table}[h]
    \footnotesize
    \centering
    \begin{tabular}{@{}lcc@{}}
        \toprule
        {Comparison} & {$p$} & {Decision}  \\ 
        \midrule
        Context vs.\ No-Context   & $0.000$ & Reject $H_0$  \\
        Comment vs.\ No-Comment    & $0.032$ & Fail to reject $H_{0}$          \\
        Markdown vs.\ No-Markdown   & $0.054$ & Fail to reject $H_{0}$        \\
        \bottomrule
    \end{tabular}
    \caption{Wilcoxon signed-rank test results ($\alpha = 0.01$) - $p$-values rounded to 3-decimals. }
    \label{tbl:wilcoxon-ablation}
\end{table}

Therefore, we conclude that \emph{context} helps,
however, the results of \emph{comment} and \emph{markdown} are inconclusive,
as we do not find statistical evidence to reject the respective $H_0$.

\noindent
\textbf{Formal Languages}.
Recall $n=80$ for each formal language $l \in \{\text{Pyke},\text{ASP},\text{NLTK},\text{FOL}\}$.
For comparing the formal languages, we perform a two-step approach:
First, we perform the Friedman test to determine whether there is a difference between (Pyke,ASP,NLTK,FOL),
with $H_0$ that there is no difference.
If we reject Friedman's $H_0$, we perform $6$ pairwise Wilcoxon signed rank tests: (FOL,NLTK), (FOL,ASP), (FOL,Pyke), (NLTK,ASP), (NLTK,Pyke), (ASP,Pyke).
For the Friedman test we obtain a $p$-value of $p=0.00$,
therefore we reject $H_0$.
We go on to perform the pairwise tests,
with the Wilcoxon signed rank test - $H_0$ being \emph{there is no difference}
and $H_1$ being \emph{language-1} $>$ \emph{language-2}.
We report the $p$-value and the Holm corrected $p$-value in Table~\ref{tbl:wilcoxon-language}.
%
%

\begin{table}[h]
    \footnotesize
    \centering
    \begin{tabular}{@{}lccc@{}}
        \toprule
        {Comparison} & {Raw $p$} & {Holm-adj. $p$} & {Decision}  \\ 
        \midrule
        FOL vs.\ NLTK   & $0.011$ & $0.022$ & Fail to reject $H_{0}$  \\
        FOL vs.\ ASP    & $0.000$ & $0.000$ & Reject $H_{0}$          \\
        FOL vs.\ Pyke   & $0.000$ & $0.000$ & Reject $H_{0}$        \\
        NLTK vs.\ ASP   & $0.012$ & $0.022$ & Fail to reject $H_{0}$  \\
        NLTK vs.\ Pyke  & $0.000$ & $0.000$ & Reject $H_{0}$        \\
        ASP vs.\ Pyke   & $0.000$ & $0.000$ & Reject $H_{0}$        \\
        \bottomrule
    \end{tabular}
    \caption{Wilcoxon signed-rank test results ($\alpha = 0.01$) - $p$-values rounded to 3-decimals. }
    \label{tbl:wilcoxon-language}
\end{table}

We conclude that FOL performs better than ASP and Pyke.
NLTK and ASP perform better than Pyke.
However, the results between FOL and NLTK, and NLTK and ASP, are inconclusive,
as we cannot reject the respective $H_0$.

\begin{table*}[t]
    \centering
    \resizebox{14cm}{!}{
    \centering
    \begin{tabular}{ll|ccc|ccc|ccc}
    \toprule
        \multicolumn{2}{c}{Method} & \multicolumn{3}{c}{ProntoQA} & \multicolumn{3}{c}{ProofWriter} & \multicolumn{3}{c}{FOLIO} \\
        \cmidrule(lr){3-5} \cmidrule(lr){6-8} \cmidrule(lr){9-11}   &  & Overall-Acc & Exec-Rate & Exec-Acc & Overall-Acc & Exec-Rate & Exec-Acc & Overall-Acc & Exec-Rate & Exec-Acc\\
    \midrule
Chance &  & 50.00 & / & / & 33.33 & / & / & 33.33 & / & / \\ \midrule
\multirow{4}{*}{Baseline} & Standard & 70.20 & 100.00 & 70.20 & 53.50 & 100.00 & 53.50 & 63.24 & 100.00 & 63.24 \\
 & CoT & 84.00 & 100.00 & 84.00 & 49.33 & 100.00 & 49.33 & \textbf{66.18} & 100.00 & 66.18 \\
 & Logic-LM & 74.40 & 100.00 & 74.40 & 0.00 & 0.00 & 0.00 & 0.00 & 0.00 & 0.00 \\
 & LINC* & 43.60 & 43.60 & 100.00 & 36.67 & 39.33 & 93.22 & 21.57 & 33.33 & 64.71 \\ \midrule
\multirow{8}{*}{Pyke} & No-C. & 41.60 & 82.00 & 50.73 & 38.67 & 74.00 & 52.25 & / & / & / \\
 & Text & 75.40 & 99.00 & 76.16 & 45.83 & 62.83 & 72.94 & / & / & / \\
 & Comm.-No-C. & 54.00 & 97.20 & 55.56 & 47.50 & 89.00 & 53.37 & / & / & / \\
 & Comm.-Text & 86.00 & 99.20 & 86.69 & 41.17 & 58.33 & 70.57 & / & / & / \\
 & MD-No-C. & 49.80 & 97.60 & 51.02 & 49.83 & 83.50 & 59.68 & / & / & / \\
 & MD-Text & 93.80 & 99.80 & 93.99 & 56.33 & 75.67 & 74.45 & / & / & / \\
 & MD-Comm.-No-C. & 71.40 & 100.00 & 71.40 & 52.83 & 71.67 & 73.72 & / & / & / \\
 & MD-Comm.-Text & 91.60 & 99.00 & 92.53 & 60.33 & 78.83 & 76.53 & / & / & / \\ \midrule
\multirow{8}{*}{ASP} & No-C. & 42.40 & 86.00 & 49.30 & 49.17 & 95.33 & 51.57 & / & / & / \\
 & Text & 70.40 & 90.40 & 77.88 & 54.83 & 82.17 & 66.73 & / & / & / \\
 & Comm.-No-C. & 61.20 & 100.00 & 61.20 & 70.00 & 98.67 & 70.95 & / & / & / \\
 & Comm.-Text & 50.00 & 100.00 & 50.00 & 33.33 & 100.00 & 33.33 & / & / & / \\
 & MD-No-C. & 43.60 & 90.20 & 48.34 & 49.67 & 99.67 & 49.83 & / & / & / \\
 & MD-Text & 81.40 & 92.20 & 88.29 & 59.83 & 92.33 & 64.80 & / & / & / \\
 & MD-Comm.-No-C. & 62.80 & 100.00 & 62.80 & 79.67 & 100.00 & 79.67 & / & / & / \\
 & MD-Comm.-Text & 97.20 & 97.40 & 99.79 & 70.50 & 91.00 & 77.47 & / & / & / \\ \midrule\
\multirow{8}{*}{NLTK} & No-C. & 48.80 & 99.80 & 48.90 & 56.17 & 97.17 & 57.80 & 40.43 & 84.04 & 48.10 \\
 & Text & 95.20 & 100.00 & 95.20 & 81.33 & 88.33 & 92.08 & 48.33 & 78.33 & 61.70 \\
 & Comm.-No-C. & 48.80 & 99.80 & 48.90 & 56.17 & 97.17 & 57.80 & 58.82 & 94.61 & 62.18 \\
 & Comm.-Text & 99.80 & 100.00 & 99.80 & 95.67 & 98.67 & 96.96 & 63.33 & 83.33 & 76.00 \\
 & MD-No-C. & 48.80 & 99.80 & 48.90 & 56.17 & 97.17 & 57.80 & 44.12 & 92.65 & 47.62 \\
 & MD-Text & 95.20 & 100.00 & 95.20 & 81.33 & 88.33 & 92.08 & 48.53 & 77.94 & 62.26 \\
 & MD-Comm.-No-C. & 48.80 & 99.80 & 48.90 & 56.17 & 97.17 & 57.80 & 59.80 & 90.69 & 65.95 \\
 & MD-Comm.-Text & 99.80 & 100.00 & 99.80 & 95.67 & 98.67 & 96.96 & 54.90 & 85.29 & 64.37 \\ \midrule
\multirow{8}{*}{FOL} & No-C. & 49.60 & 99.60 & 49.80 & 62.33 & 94.33 & 66.08 & 40.38 & 86.54 & 46.67 \\
 & Text & 90.00 & 100.00 & 90.00 & 79.83 & 85.67 & 93.19 & 52.38 & 77.38 & 67.69 \\
 & Comm.-No-C. & 49.60 & 99.60 & 49.80 & 62.33 & 94.33 & 66.08 & 60.29 & 91.67 & 65.78 \\
 & Comm.-Text & \textbf{100.00} & 100.00 & 100.00 & \textbf{97.00} & 99.83 & 97.16 & 50.00 & 70.00 & 71.43 \\
 & MD-No-C. & 49.60 & 99.60 & 49.80 & 62.33 & 94.33 & 66.08 & 0.00 & 0.00 & 0.00 \\
 & MD-Text & 90.00 & 100.00 & 90.00 & 79.83 & 85.67 & 93.19 & 48.04 & 81.37 & 59.04 \\
 & MD-Comm.-No-C. & 49.60 & 99.60 & 49.80 & 62.33 & 94.33 & 66.08 & 61.76 & 93.63 & 65.97 \\
 & MD-Comm.-Text & \textbf{100.00} & 100.00 & 100.00 & 81.17 & 86.50 & 93.83 & 55.88 & 79.90 & 69.94 \\
    \bottomrule
    \end{tabular}
        }
        \caption{
        Detailed results for the GPT-4o-mini model, depicting overall-accuracy, execution-rate, and execution-accuracy for the ProntoQA, ProofWriter, and FOLIO datasets.
        All values shown in percent [\%].
        }
        \label{tbl:asp-table-our-results}
\end{table*}

\begin{table*}[t]
    \centering
    \resizebox{14cm}{!}{
    \centering
    \begin{tabular}{ll|ccc|ccc|ccc}
    \toprule
        \multicolumn{2}{c}{Method} & \multicolumn{3}{c}{ProntoQA} & \multicolumn{3}{c}{ProofWriter} & \multicolumn{3}{c}{FOLIO}\\

        \cmidrule(lr){3-5} \cmidrule(lr){6-8} \cmidrule(lr){9-11}   &  & Overall-Acc & Exec-Rate & Exec-Acc & Overall-Acc & Exec-Rate & Exec-Acc & Overall-Acc & Exec-Rate & Exec-Acc\\
    \midrule
Chance &  & 50.00 & / & / & 33.33 & / & / & 33.33 & / & / \\ \midrule
\multirow{4}{*}{Baseline} & Standard & 52.00 & 100.00 & 52.00 & 43.50 & 100.00 & 43.50 & 52.94 & 100.00 & 52.94 \\
 & CoT & 68.80 & 100.00 & 68.80 & 46.67 & 100.00 & 46.67 & \textbf{57.35} & 100.00 & 57.35 \\
 & Logic-LM & 0.00 & 0.00 & 0.00 & 0.00 & 0.00 & 0.00 & 0.00 & 0.00 & 0.00 \\
 & LINC* & 15.00 & 15.00 & 100.00 & 21.67 & 24.50 & 88.44 & 0.00 & 0.00 & 0.00 \\ \midrule
\multirow{8}{*}{Pyke} & No-C. & 31.80 & 56.40 & 56.38 & 5.50 & 11.83 & 46.48 & / & / & / \\
 & Text & 55.40 & 73.60 & 75.27 & 0.00 & 0.33 & 0.00 & / & / & / \\
 & Comm.-No-C. & 27.00 & 53.00 & 50.94 & 3.67 & 6.50 & 56.41 & / & / & / \\
 & Comm.-Text & 0.00 & 0.00 & 0.00 & 7.50 & 11.00 & 68.18 & / & / & / \\
 & MD-No-C. & 24.20 & 45.80 & 52.84 & 3.17 & 10.67 & 29.69 & / & / & / \\
 & MD-Text & 61.00 & 76.00 & 80.26 & 0.17 & 0.17 & 100.00 & / & / & / \\
 & MD-Comm.-No-C. & 36.00 & 66.60 & 54.05 & 10.00 & 19.33 & 51.72 & / & / & / \\
 & MD-Comm.-Text & 57.00 & 64.80 & 87.96 & 9.83 & 12.83 & 76.62 & / & / & / \\ \midrule
\multirow{8}{*}{ASP} & No-C. & 9.60 & 21.80 & 44.04 & 1.83 & 5.33 & 34.38 & 0.00 & 0.00 & 0.00 \\
 & Text & 7.00 & 11.40 & 61.40 & 12.33 & 29.33 & 42.05 & 2.94 & 6.86 & 42.86 \\
 & Comm.-No-C. & 6.00 & 13.20 & 45.45 & 0.67 & 2.67 & 25.00 & 0.49 & 1.47 & 33.33 \\
 & Comm.-Text & 7.00 & 10.60 & 66.04 & 4.33 & 8.33 & 52.00 & 0.98 & 2.45 & 40.00 \\
 & MD-No-C. & 9.80 & 18.80 & 52.13 & 2.17 & 5.33 & 40.62 & 0.98 & 3.92 & 25.00 \\
 & MD-Text & 5.40 & 9.40 & 57.45 & 19.33 & 38.00 & 50.88 & 7.35 & 22.55 & 32.61 \\
 & MD-Comm.-No-C. & 7.40 & 13.60 & 54.41 & 3.83 & 8.17 & 46.94 & 0.49 & 2.45 & 20.00 \\
 & MD-Comm.-Text & 8.20 & 12.00 & 68.33 & 3.67 & 7.83 & 46.81 & 0.00 & 0.00 & 0.00 \\ \midrule
\multirow{8}{*}{NLTK} & No-C. & 12.80 & 25.40 & 50.39 & 22.67 & 40.50 & 55.97 & 2.94 & 7.35 & 40.00 \\
 & Text & 32.00 & 34.80 & 91.95 & 35.67 & 41.50 & 85.94 & 4.41 & 10.29 & 42.86 \\
 & Comm.-No-C. & 32.80 & 46.00 & 71.30 & 28.00 & 33.33 & 84.00 & 34.80 & 65.69 & 52.99 \\
 & Comm.-Text & 34.80 & 36.60 & 95.08 & 38.67 & 41.67 & 92.80 & 27.45 & 50.98 & 53.85 \\
 & MD-No-C. & 11.60 & 23.20 & 50.00 & 25.33 & 49.50 & 51.18 & 7.84 & 17.16 & 45.71 \\
 & MD-Text & 61.40 & 66.20 & 92.75 & 45.50 & 53.83 & 84.52 & 17.65 & 39.71 & 44.44 \\
 & MD-Comm.-No-C. & 48.80 & 69.20 & 70.52 & 33.67 & 42.67 & 78.91 & 40.20 & 70.10 & 57.34 \\
 & MD-Comm.-Text & 55.20 & 60.00 & 92.00 & 40.83 & 45.83 & 89.09 & 41.67 & 57.35 & 72.65 \\ \midrule
\multirow{8}{*}{FOL} & No-C. & 26.40 & 50.00 & 52.80 & 28.17 & 57.33 & 49.13 & 0.00 & 1.47 & 0.00 \\
 & Text & 70.20 & 80.60 & 87.10 & 64.33 & 80.83 & 79.59 & 6.37 & 9.31 & 68.42 \\
 & Comm.-No-C. & 59.00 & 67.80 & 87.02 & 45.33 & 57.00 & 79.53 & 23.53 & 42.16 & 55.81 \\
 & Comm.-Text & 42.60 & 52.00 & 81.92 & \textbf{66.33} & 85.33 & 77.73 & 32.84 & 51.96 & 63.21 \\
 & MD-No-C. & 19.80 & 36.80 & 53.80 & 29.00 & 61.00 & 47.54 & 0.49 & 0.98 & 50.00 \\
 & MD-Text & \textbf{77.00} & 83.40 & 92.33 & 63.17 & 79.50 & 79.45 & 16.67 & 32.84 & 50.75 \\
 & MD-Comm.-No-C. & 73.40 & 80.40 & 91.29 & 40.83 & 52.50 & 77.78 & 26.96 & 46.08 & 58.51 \\
 & MD-Comm.-Text & 72.20 & 77.40 & 93.28 & 53.83 & 61.33 & 87.77 & 38.24 & 55.88 & 68.42 \\
    \bottomrule
    \end{tabular}
        }
        \caption{
        Detailed results for the Llama 3.1 8B Instruct model, depicting overall-accuracy, execution-rate, and execution-accuracy for the ProntoQA, ProofWriter, and FOLIO datasets.
        All values shown in percent [\%].
        }
        \label{tbl:llama-8b-results}
\end{table*}

\begin{table*}[t]
    \centering
    \resizebox{14cm}{!}{
    \centering
    \begin{tabular}{ll|ccc|ccc|ccc}
    \toprule
        \multicolumn{2}{c}{Method} & \multicolumn{3}{c}{ProntoQA} & \multicolumn{3}{c}{ProofWriter} & \multicolumn{3}{c}{FOLIO} \\

        \cmidrule(lr){3-5} \cmidrule(lr){6-8} \cmidrule(lr){9-11}           &  & Overall-Acc & Exec-Rate & Exec-Acc & Overall-Acc & Exec-Rate & Exec-Acc & Overall-Acc & Exec-Rate & Exec-Acc\\
    \midrule
Chance &  & 50.00 & / & / & 33.33 & / & / & 33.33 & / & / \\ \midrule
\multirow{4}{*}{Baseline} & Standard & 52.00 & 100.00 & 52.00 & 43.50 & 100.00 & 43.50 & 52.94 & 100.00 & 52.94 \\
 & CoT & 68.80 & 100.00 & 68.80 & \textbf{46.67} & 100.00 & 46.67 & \textbf{57.35} & 100.00 & 57.35 \\
 & Logic-LM & 0.00 & 0.00 & 0.00 & 0.00 & 0.00 & 0.00 & 0.00 & 0.00 & 0.00 \\
 & LINC* & 15.00 & 15.00 & 100.00 & 21.67 & 24.50 & 88.44 & 0.00 & 0.00 & 0.00 \\ \midrule
\multirow{8}{*}{Pyke} & No-C. & 4.20 & 8.00 & 52.50 & 0.33 & 0.83 & 40.00 & / & / & / \\
 & Text & 5.80 & 13.40 & 43.28 & 0.50 & 0.67 & 75.00 & / & / & / \\
 & Comm.-No-C. & 0.40 & 1.20 & 33.33 & 0.17 & 0.33 & 50.00 & / & / & / \\
 & Comm.-Text & 0.20 & 0.80 & 25.00 & 0.00 & 0.00 & 0.00 & / & / & / \\
 & MD-No-C. & 2.00 & 4.60 & 43.48 & 0.17 & 0.33 & 50.00 & / & / & / \\
 & MD-Text & 5.00 & 9.80 & 51.02 & 0.67 & 1.17 & 57.14 & / & / & / \\
 & MD-Comm.-No-C. & 1.80 & 3.00 & 60.00 & 0.00 & 0.00 & 0.00 & / & / & / \\
 & MD-Comm.-Text & 1.60 & 2.20 & 72.73 & 0.00 & 0.00 & 0.00 & / & / & / \\ \midrule
\multirow{8}{*}{ASP} & No-C. & 9.60 & 21.80 & 44.04 & 1.83 & 5.33 & 34.38 & / & / & / \\
 & Text & 7.00 & 11.40 & 61.40 & 12.33 & 29.33 & 42.05 & / & / & / \\
 & Comm.-No-C. & 6.00 & 13.20 & 45.45 & 0.67 & 2.67 & 25.00 & / & / & / \\
 & Comm.-Text & 7.00 & 10.60 & 66.04 & 4.33 & 8.33 & 52.00 & / & / & / \\
 & MD-No-C. & 9.80 & 18.80 & 52.13 & 2.17 & 5.33 & 40.62 & / & / & / \\
 & MD-Text & 5.40 & 9.40 & 57.45 & 19.33 & 38.00 & 50.88 & / & / & / \\
 & MD-Comm.-No-C. & 7.40 & 13.60 & 54.41 & 3.83 & 8.17 & 46.94 & / & / & / \\
 & MD-Comm.-Text & 8.20 & 12.00 & 68.33 & 3.67 & 7.83 & 46.81 & / & / & / \\ \midrule
\multirow{8}{*}{NLTK} & No-C. & 19.60 & 40.00 & 49.00 & 18.67 & 33.50 & 55.72 & 3.43 & 14.71 & 23.33 \\
 & Text & 18.60 & 30.40 & 61.18 & 23.17 & 35.83 & 64.65 & 1.96 & 5.39 & 36.36 \\
 & Comm.-No-C. & 15.60 & 26.20 & 59.54 & 7.17 & 11.67 & 61.43 & 1.96 & 3.43 & 57.14 \\
 & Comm.-Text & 24.00 & 32.20 & 74.53 & 9.67 & 11.00 & 87.88 & 0.00 & 0.49 & 0.00 \\
 & MD-No-C. & 17.00 & 30.60 & 55.56 & 3.67 & 7.17 & 51.16 & 10.78 & 29.90 & 36.07 \\
 & MD-Text & 22.80 & 39.20 & 58.16 & 4.33 & 7.17 & 60.47 & 8.82 & 25.98 & 33.96 \\
 & MD-Comm.-No-C. & 29.00 & 49.80 & 58.23 & 9.67 & 13.67 & 70.73 & 1.96 & 4.90 & 40.00 \\
 & MD-Comm.-Text & 26.60 & 38.40 & 69.27 & 2.83 & 3.67 & 77.27 & 5.88 & 12.25 & 48.00 \\ \midrule
\multirow{8}{*}{FOL} & No-C. & 23.60 & 44.20 & 53.39 & 24.67 & 58.17 & 42.41 & 1.96 & 6.37 & 30.77 \\
 & Text & 35.20 & 56.80 & 61.97 & 22.83 & 56.67 & 40.29 & 1.96 & 6.37 & 30.77 \\
 & Comm.-No-C. & 56.00 & 83.60 & 66.99 & 29.83 & 40.50 & 73.66 & 2.45 & 7.35 & 33.33 \\
 & Comm.-Text & \textbf{69.80} & 93.40 & 74.73 & 22.50 & 57.67 & 39.02 & 1.96 & 6.37 & 30.77 \\
 & MD-No-C. & 16.60 & 32.20 & 51.55 & 17.00 & 48.00 & 35.42 & 2.94 & 2.94 & 100.00 \\
 & MD-Text & 39.80 & 63.40 & 62.78 & 20.33 & 47.83 & 42.51 & 2.94 & 4.90 & 60.00 \\
 & MD-Comm.-No-C. & 54.20 & 81.40 & 66.58 & 22.67 & 33.17 & 68.34 & 5.39 & 8.33 & 64.71 \\
 & MD-Comm.-Text & 64.40 & 86.60 & 74.36 & 11.00 & 15.33 & 71.74 & 6.37 & 10.78 & 59.09 \\
\midrule
    \bottomrule
    \end{tabular}
        }
        \caption{
        Detailed results for the DeepSeek-8b model, depicting overall-accuracy, execution-rate, and execution-accuracy for the ProntoQA, ProofWriter, and FOLIO datasets.
        All values shown in percent [\%].
        }
        \label{tbl:deepseek-8b}
\end{table*}

\begin{table*}[t]
    \centering
    \resizebox{14cm}{!}{
    \centering
    \begin{tabular}{ll|ccc|ccc|ccc}
    \toprule
        \multicolumn{2}{c}{Method} & \multicolumn{3}{c}{ProntoQA} & \multicolumn{3}{c}{ProofWriter} & \multicolumn{3}{c}{FOLIO} \\

        \cmidrule(lr){3-5} \cmidrule(lr){6-8} \cmidrule(lr){9-11}           &  & Overall-Acc & Exec-Rate & Exec-Acc & Overall-Acc & Exec-Rate & Exec-Acc & Overall-Acc & Exec-Rate & Exec-Acc\\
    \midrule
Chance &  & 50.00 & / & / & 33.33 & / & / & 33.33 & / & / \\ \midrule
\multirow{4}{*}{Baseline} & Standard & \textbf{99.20} & 100.00 & 99.20 & 64.17 & 100.00 & 64.17 & 67.16 & 100.00 & 67.16 \\
 & CoT & 98.80 & 100.00 & 98.80 & 66.67 & 100.00 & 66.67 & \textbf{71.57} & 100.00 & 71.57 \\
 & Logic-LM & 74.20 & 88.00 & 84.32 & 0.00 & 0.00 & 0.00 & 0.00 & 0.00 & 0.00 \\
 & LINC* & 0.00 & 0.00 & 0.00 & 0.00 & 0.17 & 0.00 & 0.00 & 0.00 & 0.00 \\ \midrule
\multirow{8}{*}{Pyke} & No-C. & 42.40 & 64.60 & 65.63 & 56.17 & 80.83 & 69.48 & / & / & / \\
 & Text & 74.60 & 89.00 & 83.82 & 46.67 & 54.17 & 86.15 & / & / & / \\
 & Comm.-No-C. & 43.00 & 62.60 & 68.69 & 38.17 & 56.17 & 67.95 & / & / & / \\
 & Comm.-Text & 66.40 & 81.60 & 81.37 & 37.33 & 44.83 & 83.27 & / & / & / \\
 & MD-No-C. & 43.00 & 70.00 & 61.43 & 54.50 & 68.00 & 80.15 & / & / & / \\
 & MD-Text & 68.80 & 83.20 & 82.69 & 45.67 & 51.67 & 88.39 & / & / & / \\
 & MD-Comm.-No-C. & 68.80 & 83.00 & 82.89 & 50.83 & 71.50 & 71.10 & / & / & / \\
 & MD-Comm.-Text & 79.40 & 89.80 & 88.42 & 39.33 & 46.83 & 83.99 & / & / & / \\ \midrule
\multirow{8}{*}{ASP} & No-C. & 63.60 & 87.00 & 73.10 & 50.50 & 74.00 & 68.24 & / & / & / \\
 & Text & 72.00 & 90.00 & 80.00 & 66.00 & 85.17 & 77.50 & / & / & / \\
 & Comm.-No-C. & 72.00 & 95.20 & 75.63 & 62.83 & 80.67 & 77.89 & / & / & / \\
 & Comm.-Text & 67.80 & 94.40 & 71.82 & 60.33 & 75.67 & 79.74 & / & / & / \\
 & MD-No-C. & 73.20 & 94.60 & 77.38 & 54.17 & 76.67 & 70.65 & / & / & / \\
 & MD-Text & 77.60 & 92.40 & 83.98 & 65.33 & 84.33 & 77.47 & / & / & / \\
 & MD-Comm.-No-C. & 52.20 & 93.00 & 56.13 & 64.83 & 80.67 & 80.37 & / & / & / \\
 & MD-Comm.-Text & 58.60 & 95.80 & 61.17 & 58.83 & 74.67 & 78.79 & / & / & / \\ \midrule
\multirow{8}{*}{NLTK} & No-C. & 35.60 & 50.40 & 70.63 & 67.00 & 79.67 & 84.10 & 42.16 & 57.35 & 73.50 \\
 & Text & 96.00 & 97.80 & 98.16 & 74.50 & 83.00 & 89.76 & 51.96 & 68.63 & 75.71 \\
 & Comm.-No-C. & 56.20 & 60.20 & 93.36 & 54.33 & 64.00 & 84.90 & 4.41 & 7.84 & 56.25 \\
 & Comm.-Text & 67.60 & 68.00 & 99.41 & 45.67 & 50.83 & 89.84 & 6.37 & 7.84 & 81.25 \\
 & MD-No-C. & 40.40 & 66.40 & 60.84 & 68.33 & 79.17 & 86.32 & 26.96 & 36.76 & 73.33 \\
 & MD-Text & 95.00 & 96.80 & 98.14 & 72.83 & 83.00 & 87.75 & 48.04 & 64.71 & 74.24 \\
 & MD-Comm.-No-C. & 59.60 & 66.20 & 90.03 & 23.33 & 31.67 & 73.68 & 2.94 & 6.37 & 46.15 \\
 & MD-Comm.-Text & 74.80 & 76.40 & 97.91 & 19.83 & 27.17 & 73.01 & 1.96 & 4.41 & 44.44 \\ \midrule
\multirow{8}{*}{FOL} & No-C. & 49.20 & 57.80 & 85.12 & 75.67 & 85.50 & 88.50 & 49.02 & 65.20 & 75.19 \\
 & Text & 85.00 & 89.40 & 95.08 & 74.83 & 83.00 & 90.16 & 43.63 & 60.78 & 71.77 \\
 & Comm.-No-C. & 61.60 & 67.00 & 91.94 & 50.83 & 61.67 & 82.43 & 12.75 & 16.67 & 76.47 \\
 & Comm.-Text & 55.20 & 56.80 & 97.18 & \textbf{80.00} & 87.33 & 91.60 & 4.90 & 11.76 & 41.67 \\
 & MD-No-C. & 48.60 & 58.80 & 82.65 & 78.67 & 87.17 & 90.25 & 43.63 & 54.90 & 79.46 \\
 & MD-Text & 87.00 & 90.40 & 96.24 & 75.67 & 83.50 & 90.62 & 45.10 & 59.80 & 75.41 \\
 & MD-Comm.-No-C. & 52.20 & 60.00 & 87.00 & 14.00 & 28.00 & 50.00 & 5.88 & 8.33 & 70.59 \\
 & MD-Comm.-Text & 75.60 & 79.40 & 95.21 & 13.67 & 30.17 & 45.30 & 3.92 & 8.82 & 44.44 \\
    \bottomrule
    \end{tabular}
        }
        \caption{
        Detailed results for the DeepSeek-32b model, depicting overall-accuracy, execution-rate, and execution-accuracy for the ProntoQA, ProofWriter, and FOLIO datasets.
        All values shown in percent [\%].
        }
        \label{tbl:deepseek-32b}
\end{table*}

\begin{table*}[t]
    \centering
    \resizebox{14cm}{!}{
    \centering
    \begin{tabular}{ll|ccc|ccc|ccc}
    \toprule
        \multicolumn{2}{c}{Method} & \multicolumn{3}{c}{ProntoQA} & \multicolumn{3}{c}{ProofWriter} & \multicolumn{3}{c}{FOLIO} \\

        \cmidrule(lr){3-5} \cmidrule(lr){6-8} \cmidrule(lr){9-11}           &  & Overall-Acc & Exec-Rate & Exec-Acc & Overall-Acc & Exec-Rate & Exec-Acc & Overall-Acc & Exec-Rate & Exec-Acc\\
Chance &  & 50.00 & / & / & 33.33 & / & / & 33.33 & / & / \\ \midrule
\multirow{4}{*}{Baseline} & Standard & 98.00 & 100.00 & 98.00 & 76.00 & 100.00 & 76.00 & 68.63 & 100.00 & 68.63 \\
 & CoT & 99.80 & 100.00 & 99.80 & 79.67 & 100.00 & 79.67 & \textbf{71.57} & 100.00 & 71.57 \\
 & Logic-LM & 73.60 & 99.60 & 73.90 & 0.00 & 0.00 & 0.00 & 0.00 & 0.00 & 0.00 \\
 & LINC* & 0.00 & 0.00 & 0.00 & 0.00 & 0.00 & 0.00 & 0.00 & 0.00 & 0.00 \\ \midrule
\multirow{8}{*}{Pyke} & No-C. & 49.60 & 83.00 & 59.76 & 49.00 & 67.67 & 72.41 & / & / & / \\
 & Text & 78.20 & 100.00 & 78.20 & 77.00 & 96.67 & 79.66 & / & / & / \\
 & Comm.-No-C. & 56.80 & 86.20 & 65.89 & 66.50 & 93.83 & 70.87 & / & / & / \\
 & Comm.-Text & 74.00 & 99.40 & 74.45 & 72.33 & 87.67 & 82.51 & / & / & / \\
 & MD-No-C. & 47.00 & 83.60 & 56.22 & 73.17 & 94.17 & 77.70 & / & / & / \\
 & MD-Text & 78.40 & 100.00 & 78.40 & 77.83 & 96.83 & 80.38 & / & / & / \\
 & MD-Comm.-No-C. & 78.60 & 95.40 & 82.39 & 64.67 & 81.00 & 79.84 & / & / & / \\
 & MD-Comm.-Text & 82.40 & 99.80 & 82.57 & 65.67 & 81.00 & 81.07 & / & / & / \\ \midrule
\multirow{8}{*}{ASP} & No-C. & 30.60 & 55.80 & 54.84 & 74.67 & 98.83 & 75.55 & / & / & / \\
 & Text & 92.60 & 100.00 & 92.60 & 78.83 & 100.00 & 78.83 & / & / & / \\
 & Comm.-No-C. & 95.60 & 100.00 & 95.60 & 79.83 & 100.00 & 79.83 & / & / & / \\
 & Comm.-Text & 98.80 & 100.00 & 98.80 & 79.83 & 100.00 & 79.83 & / & / & / \\
 & MD-No-C. & 17.40 & 32.40 & 53.70 & 74.00 & 99.17 & 74.62 & / & / & / \\
 & MD-Text & 99.00 & 100.00 & 99.00 & 78.67 & 99.67 & 78.93 & / & / & / \\
 & MD-Comm.-No-C. & 94.60 & 100.00 & 94.60 & 79.83 & 100.00 & 79.83 & / & / & / \\
 & MD-Comm.-Text & 99.00 & 100.00 & 99.00 & 79.83 & 100.00 & 79.83 & / & / & / \\ \midrule
\multirow{8}{*}{NLTK} & No-C. & 34.80 & 61.80 & 56.31 & 93.17 & 96.83 & 96.21 & 63.73 & 85.29 & 74.71 \\
 & Text & 99.20 & 100.00 & 99.20 & 91.33 & 94.00 & 97.16 & 57.84 & 80.88 & 71.52 \\
 & Comm.-No-C. & 99.40 & 99.60 & 99.80 & 79.67 & 82.83 & 96.18 & 57.35 & 75.49 & 75.97 \\
 & Comm.-Text & 96.40 & 96.40 & 100.00 & 71.33 & 73.17 & 97.49 & 49.51 & 67.16 & 73.72 \\
 & MD-No-C. & 32.20 & 58.20 & 55.33 & 94.17 & 97.67 & 96.42 & 64.71 & 87.25 & 74.16 \\
 & MD-Text & \textbf{100.00} & 100.00 & 100.00 & 92.17 & 94.83 & 97.19 & 61.76 & 80.88 & 76.36 \\
 & MD-Comm.-No-C. & 99.40 & 100.00 & 99.40 & 82.67 & 85.67 & 96.50 & 61.27 & 77.94 & 78.62 \\
 & MD-Comm.-Text & \textbf{100.00} & 100.00 & 100.00 & 79.17 & 81.67 & 96.94 & 33.33 & 40.69 & 81.93 \\ \midrule
\multirow{8}{*}{FOL} & No-C. & 46.20 & 76.80 & 60.16 & \textbf{94.33} & 97.00 & 97.25 & 57.84 & 82.84 & 69.82 \\
 & Text & 99.20 & 100.00 & 99.20 & 90.00 & 92.83 & 96.95 & 60.29 & 77.45 & 77.85 \\
 & Comm.-No-C. & 50.80 & 50.80 & 100.00 & 80.50 & 83.00 & 96.99 & 63.73 & 84.80 & 75.14 \\
 & Comm.-Text & 4.00 & 4.00 & 100.00 & 89.33 & 92.33 & 96.75 & 40.20 & 48.53 & 82.83 \\
 & MD-No-C. & 47.20 & 80.20 & 58.85 & 91.33 & 94.17 & 96.99 & 59.80 & 79.90 & 74.85 \\
 & MD-Text & 98.80 & 100.00 & 98.80 & 85.50 & 88.50 & 96.61 & 56.37 & 75.00 & 75.16 \\
 & MD-Comm.-No-C. & \textbf{100.00} & 100.00 & 100.00 & 84.33 & 87.17 & 96.75 & 41.18 & 54.41 & 75.68 \\
 & MD-Comm.-Text & 83.40 & 83.40 & 100.00 & 79.67 & 81.67 & 97.55 & 50.98 & 67.65 & 75.36 \\
\bottomrule
    \end{tabular}
        }
        \caption{
        DeepSeek-V3
        Detailed results for the DeepSeek-V3 model, depicting overall-accuracy, execution-rate, and execution-accuracy for the ProntoQA, ProofWriter, and FOLIO datasets.
        All values shown in percent [\%].
        }
        \label{tbl:deepseek-v3}
\end{table*}

\subsection{Qualitative Error Analysis Details}
\label{sec:app:qualitative-details}

Here, we provide details and additional common errors we found, including where they occurred.
We structure their place of occurrence as $(\langle \text{LLM} \rangle,$$\langle \text{Text/No-Context (No-C.)} \rangle ,$\\$ \langle \text{comment/no comment} \rangle ,$$ \langle \text{markdown/no markdown} \rangle ,$\\$\text{ID: } \langle \text{Example ID} \rangle)$.

\textbf{Pyke 1}:
Semantically incorrect, but syntactically correct translation:
\textit{Stella is a yumpus} is translated as \textit{Tumpus(Stella, True)} (GPT-4o-mini, text, no comment, no markdown, ID: ProntoQA\_2).\\
\textbf{Pyke 2}:
Endless output generation, which was capped at 2048 output tokens (Ministral-8b, text, no comment, markdown, ID: ProofWriter\_AttNeg-OWA-D5-1220\_Q6).
\\
\textbf{Pyke 3}:
Syntactically incorrect output:
Line breaks and tabs were missed as \textit{foreach facts.P1(x, True) assert facts.P9(x, True)} (DeepSeek-8B, No-C., comment, no markdown, ID: ProntoQA\_1).
\\
\textbf{ASP 1}:
Syntactically incorrect output:
Not adhering to task description, by producing a CoT reasoning chain (DeepSeek-32B, text, comment, no markdown, ID: ProntoQA\_3).
%
\\
\textbf{ASP 2}:
Syntactically incorrect output:
Unable to capture negation, such as the query \textit{-not p1(wren)} (Llama-8b, No-C., no comment, no markdown, ID: ProntoQA\_9).\\
%
\textbf{NLTK 1}: 
Syntactically incorrect output:
Forgot keyword introducing an NLTK formula (DeepSeek-V3, text, comment, markdown, ID: FOLIO\_dev\_1).\\
%
\textbf{NLTK 2}:
Syntactically incorrect output:
Misplaced parentheses, such as \textit{attend(x) \& engage(x) | -attend(x) \& -engage(x))} (DeepSeek-8b, text, no comment, markdown, ID: FOLIO\_dev\_0).\\
%
\textbf{FOL 1}:
Syntactically incorrect output:
Multiple arities for one predicate, such as predicate $p14$ for arities $1$ and $2$ (GPT-4o-mini, No-C., no comment, no markdown, ID: ProofWriter\_RelNeg-OWA-D5-1029\_Q2).\\
\textbf{Baselines}:
The neurosymbolic baselines were particularly prone to small syntax errors, like wrapping lines or predicates in markdown bold faced letters,
or enumerating lines,
such as enumerating and wrapping a predicate: \textit{1. **Cold(\$x, bool)**} (GPT-4o-mini, Logic-LM*, ID: ProofWriter\_AttNoneg-OWA-D5-1041\_Q1).
%

\subsection{Standard Prompt}
\label{sec:app:standard-prompt}

We show the full standard prompt, including macros which are not shown in the subsequently shown examples.
\begin{lstlisting}
Given a problem statement as contexts, the task is to answer a logical reasoning question. 
------
Context:
Each jompus is fruity. Every jompus is a wumpus. Every wumpus is not transparent. Wumpuses are tumpuses. Tumpuses are mean. Tumpuses are vumpuses. Every vumpus is cold. Each vumpus is a yumpus. Yumpuses are orange. Yumpuses are numpuses. Numpuses are dull. Each numpus is a dumpus. Every dumpus is not shy. Impuses are shy. Dumpuses are rompuses. Each rompus is liquid. Rompuses are zumpuses. Alex is a tumpus.
Question:
Question: Is the following statement true or false? Alex is not shy.

Options:
A) True
B) False

The correct option is: A
------
Context:
[[CONTEXT]]

Question: [[QUESTION]]

Options:
[[OPTIONS]]

The correct option is:
\end{lstlisting}

\subsection{Chain-of-Thought (CoT) Prompt}
\label{sec-appendix:CoT-Full-Prompt}

We show a full CoT prompt without macros.
\begin{lstlisting}
Given a problem statement as contexts, the task is to answer a logical reasoning question. 
------
Context:
Each jompus is fruity. Every jompus is a wumpus. Every wumpus is not transparent. Wumpuses are tumpuses. Tumpuses are mean. Tumpuses are vumpuses. Every vumpus is cold. Each vumpus is a yumpus. Yumpuses are orange. Yumpuses are numpuses. Numpuses are dull. Each numpus is a dumpus. Every dumpus is not shy. Impuses are shy. Dumpuses are rompuses. Each rompus is liquid. Rompuses are zumpuses. Alex is a tumpus.
Question:
Question: Is the following statement true or false? Alex is not shy.

Options:
A) True
B) False

Reasoning:
Alex is a tumpus.  Tumpuses are vumpuses. So Alex is a vumpus. Each vumpus is a yumpus. So Alex is a yumpus. Yumpuses are numpuses. So Alex is a numpus. Each numpus is a dumpus. So Alex is a dumpus. Every dumpus is not shy. So Alex is not shy.

The correct option is: A
------
[...]
\end{lstlisting}

\subsection{Logic-LM*: Example Prompt}
\label{sec-appendix:logic-lm-example}

The following shows the Logic-LM* prompt without macros.
``[...]'' indicates that we skipped rules for brevity.

\begin{lstlisting}
Task Description: You are given a problem description and a question.
In general, the task is to parse the problem description and question into a a Pyke (Python Knowledge Engine) readable format.
In more detail:
1.) Define the predicates.
2) Define the facts.
3) Define the rules.
4) Define the "query". The query has to be defined according to the following example: Given the question: "True or false: Alex is not shy".
Then you should define this as "Shy(alex, false)".
The program must by syntactically correct. A correctly parsed example is given below. The output should be given in a Pyke readable format. Therefore, be sure not to use any "bullet points", or "numberings" when printing the output. Further, no special characters like ("#") must occur.
------
Problem:
Each jompus is fruity. Every jompus is a wumpus. Every wumpus is not transparent. Wumpuses are tumpuses. Tumpuses are mean. Tumpuses are vumpuses. Every vumpus is cold. Each vumpus is a yumpus. Yumpuses are orange. Yumpuses are numpuses.
Numpuses are dull. Each numpus is a dumpus. Every dumpus is not shy. Impuses are shy. Dumpuses are rompuses. Each rompus is liquid. Rompuses are zumpuses. Alex is a tumpus.
Question:
True or false: Alex is not shy.
###
Predicates:
Jompus($x, bool) ::: Does x belong to Jompus?
[...]
Facts:
Tumpuses(Alex, True)
Rules:
Jompus($x, True) >>> Fruity($x, True)
[...]
Query:
Shy(Alex, False)
------
[...]
\end{lstlisting}

\subsection{LINC*}

The following shows the Logic-LM* prompt without macros.
``[...]'' indicates that we skipped rules for brevity.

\begin{lstlisting}
The following is a first-order logic (FOL) problem.
The problem is to determine whether the conclusion follows from the premises.
The premises are given in the form of a set of first-order logic sentences.
The conclusion is given in the form of a single first-order logic sentence.
The task is to translate each of the premises and conclusions into FOL expressions, so that the expressions can be evaluated by a theorem solver to determine whether the conclusion follows from the premises.
Expressions should be adhere to the format of the Python NLTK package logic module.


<PREMISES>
All dispensable things are environment-friendly.
[...]
</PREMISES>
<CONCLUSION>
A worksheet is not dispensable.
</CONCLUSION>
<EVALUATE>
TEXT:   All dispensable things are environment-friendly.
FOL:    all x. (Dispensable(x) -> EnvironmentFriendly(x))
[...]
</EVALUATE>
[...]
\end{lstlisting}

\subsection{Pyke (No-C., comment, no markdown) Prompt}

The following shows the Pyke (No-C., comment, no markdown) prompt without macros.
``[...]'' indicates that we skipped rules for brevity.

\begin{lstlisting}
Task Description: You are given a problem description and a question.
In general, the task is to parse the problem description and question into a a Pyke (Python Knowledge Engine) readable format.
In more detail:
1) Define the facts.
2) Define the rules.
3) Define the "query". The query has to be defined according to the following example: Given the question: "True or false: Alex is not shy".
Then you should define this as "P13(alex, false)".
The program must by syntactically correct. A correctly parsed example is given below. The output should be given in a Pyke readable format. Therefore, be sure not to use any "bullet points", or "numberings" when printing the output. Further, no special characters like ("#") must occur.
------
Problem:
Each jompus is fruity. Every jompus is a wumpus. Every wumpus is not transparent. Wumpuses are tumpuses. Tumpuses are mean. Tumpuses are vumpuses. Every vumpus is cold. Each vumpus is a yumpus. Yumpuses are orange. Yumpuses are numpuses.
Numpuses are dull. Each numpus is a dumpus. Every dumpus is not shy. Impuses are shy. Dumpuses are rompuses. Each rompus is liquid. Rompuses are zumpuses. Alex is a tumpus.
Question:
True or false: Alex is not shy.
###
Facts:
# Alex is a tumpus.
P1(Alex, True)
Rules:
# Each jompus is fruity.
fact1
	foreach
		facts.P2($x, True)
	assert
		facts.P3($x, True)
[...]
Query:
# True or false: Alex is not shy.
P13(Alex,False)
[...]
\end{lstlisting}

\subsection{Pyke (Text, no comment, no markdown) Prompt}

The following shows the Pyke (Text, no comment, no markdown) prompt without macros.
``[...]'' indicates that we skipped rules for brevity.

\begin{lstlisting}
[...]
###
Facts:
Tumpus(Alex, True)
Rules:
fact1
	foreach
		facts.Jompus($x, True)
	assert
		facts.Fruity($x, True)
[...]
Query:
Shy(Alex,False)
[...]
\end{lstlisting}

\subsection{ASP (No-C., no comment, markdown) Prompt}

The following shows the ASP (No-C., no comment, markdown) prompt without macros.
``[...]'' indicates that we skipped rules for brevity.

\begin{lstlisting}
Task Description: You are given a problem description and a question.
In general, the task is to parse the problem description and question into an Answer Set Programming (ASP) program.
In more detail:
1) Define the facts.
2) Define the rules.
3) Define the "query". The query has to be defined as a (or several) literal(s).
For example: Given the question: "True or false: Alex is not shy".
Then you should define this as "-p15(alex)".
The program must by syntactically correct. A correctly parsed example is given below. The output should be given as an ASP program (logic programming). Therefore, be sure not to use any "bullet points", or "numberings" when printing the output. Further, no special characters like ("#") must occur.
------
Problem:
Each jompus is fruity. Every jompus is a wumpus. Every wumpus is not transparent. Wumpuses are tumpuses. Tumpuses are mean. Tumpuses are vumpuses. Every vumpus is cold. Each vumpus is a yumpus. Yumpuses are orange. Yumpuses are numpuses. Numpuses are dull. Each numpus is a dumpus. Every dumpus is not shy. Impuses are shy. Dumpuses are rompuses. Each rompus is liquid. Rompuses are zumpuses. Alex is a tumpus.
Question:
True or false: Alex is not shy.
###
Facts:
```
p1(alex).
```
Rules:
```
p2(X) :- p3(X).
[...]
```
Query:
```
-p15(alex).
```
------
[...]
\end{lstlisting}

\subsection{ASP (Text, comment, markdown) Prompt}

The following shows the ASP (Text, comment, markdown) prompt without macros.
``[...]'' indicates that we skipped rules for brevity.

\begin{lstlisting}
[...]
Facts:
```
% Alex is a tumpus.
tumpus(alex).
```
Rules:
```
% Each jompus is fruity.
fruity(X) :- jompus(X).
[...]
```
Query:
```
% True or false: Alex is not shy.
-shy(alex).
```
------
[...]
\end{lstlisting}

\subsection{NLTK (No-C., no comment, no markdown) Prompt}

The following shows the NLTK (No-C., no comment, no markdown) prompt without macros.
``[...]'' indicates that we skipped rules for brevity.

\begin{lstlisting}
[...]
Task Description: You are given a problem description and a question.
In general, the task is to parse the problem description and question into a a NLTK (Natural Language Toolkit) Logic format. The NLTK library is a python library.
In more detail:
1) Define the facts.
2) Define the rules.
3) Define the "query". The query has to be defined as a (or several) literal(s).
For example: Given the question: "True or false: Alex is not shy".
Then you should define this as "-p14(alex)".
The program must by syntactically correct. A correctly parsed example is given below. The output should be given in the NLTK format. Therefore, be sure not to use any "bullet points", or "numberings" when printing the output. Further, no special characters like ("#") must occur.
------
Problem:
Each jompus is fruity. Every jompus is a wumpus. Every wumpus is not transparent. Wumpuses are tumpuses. Tumpuses are mean. Tumpuses are vumpuses. Every vumpus is cold. Each vumpus is a yumpus. Yumpuses are orange. Yumpuses are numpuses. Numpuses are dull. Each numpus is a dumpus. Every dumpus is not shy. Impuses are shy. Dumpuses are rompuses. Each rompus is liquid. Rompuses are zumpuses. Alex is a tumpus.
Question:
True or false: Alex is not shy.
###
Facts:
NLTK:   p1(alex)

Rules:
NLTK:   all x. (p2(x) -> p3(x))
[...]

Query:
NLTK:   -p14(alex)
------
[...]
\end{lstlisting}

\subsection{NLTK (Text, comment, no markdown) Prompt}

The following shows the NLTK (Text, comment, no markdown) prompt without macros.
``[...]'' indicates that we skipped rules for brevity.

\begin{lstlisting}
[...]
Facts:
TEXT:   Alex is a tumpus.
NLTK:   tumpus(alex)

Rules:
TEXT:   Each jompus is fruity.
NLTK:   all x. (jompus(x) -> fruity(x))
[...]
Query:
TEXT:   Alex is not shy.
NLTK:   -shy(alex)
------
[...]
\end{lstlisting}

\subsection{FOL (No-C., comment, markdown) Prompt}

The following shows the FOL (No-C., comment, markdown) prompt without macros.
``[...]'' indicates that we skipped lines for brevity.

\begin{lstlisting}
Task Description: You are given a problem description and a question.
In general, the task is to parse the problem description and question into a a FOL (First Order Logic) format.
In more detail:
1) Define the facts.
2) Define the rules.
3) Define the "query". The query has to be defined as a (or several) literal(s).
For example: Given the question: "True or false: Alex is not shy".
Then you should define this as "-p14(alex)".
The program must by syntactically correct. A correctly parsed example is given below. The output should be given in the FOL format. Therefore, be sure not to use any "bullet points", or "numberings" when printing the output. Further, no special characters like ("#") must occur.
------
Problem:
Each jompus is fruity. Every jompus is a wumpus. Every wumpus is not transparent. Wumpuses are tumpuses. Tumpuses are mean. Tumpuses are vumpuses. Every vumpus is cold. Each vumpus is a yumpus. Yumpuses are orange. Yumpuses are numpuses. Numpuses are dull. Each numpus is a dumpus. Every dumpus is not shy. Impuses are shy. Dumpuses are rompuses. Each rompus is liquid. Rompuses are zumpuses. Alex is a tumpus.
Question:
True or false: Alex is not shy.
###
Facts:
```
TEXT:   Alex is a tumpus.
FOL:   p1(alex)
```

Rules:
```
TEXT:   Each jompus is fruity.
FOL:   %*∀*)x. (p2(x) %*→*) p3(x))
[...]
```

Query:
```
TEXT:   Alex is not shy.
FOL:   -p14(alex)
```
------
[...]
\end{lstlisting}

\subsection{FOL (Text, no comment, markdown) Prompt}
\label{sec:app:fol-prompt}

The following shows the FOL (Text, no comment, markdown) prompt without macros.
``[...]'' indicates that we skipped lines for brevity.

\begin{lstlisting}
[...]
Facts:
```
FOL:   tumpus(alex)
```

Rules:
```
FOL:   %*∀*)x. (jompus(x) %*→*) fruity(x))
[...]
```

Query:
```
FOL:   -shy(alex)
```
------
[...]
\end{lstlisting}

\subsection{Licenses of Scientific Artifacts}
\label{sec:app:licences-scientific-artifacts}

Logic-LM uses an MIT license, which grants us the free usage
and the rights to use, copy, modify, merge, publish, distribute, sublicense the software~\footnote{\url{https://github.com/teacherpeterpan/Logic-LLM}}.
ProntoQA is licensed under the Apache License 2.0~\footnote{\url{https://github.com/asaparov/prontoqa/blob/main/LICENSE}} and
FOLIO under the Creative Commons Attribution Share Alike 4.0 International~\footnote{\url{https://github.com/Yale-LILY/FOLIO/blob/main/LICENSE}},
which permits the usage of the data.
No specific license is given for LINC~\footnote{\url{https://github.com/benlipkin/linc}} and ProofWriter~\footnote{\url{https://allenai.org/data/proofwriter}}.
However, for ProofWriter they state in the paper that ``Datasets available at \url{https://allenai.org/data/proofwriter}''~\cite{tafjord_proofwriter_2021}
and for LINC ``Code is provided to reproduce all experiments and figures''~\footnote{\url{https://github.com/benlipkin/linc}}.
While we did not use any LINC code (just the prompting methodology),
we interpret the license of ProofWriter to grant us permission to use it as a benchmark dataset.
The used LLMs and software are suitable for scientific use.

\subsection{Usage of AI Assistants}
\label{sec:app:usage-of-AI-assistants}

This paper discusses the logical reasoning ability of AI assistants (LLMs).
Evidently, we performed experiments upon them - see also our experiment setup in Section~\ref{section:experimental-scenarios}.
Besides that, we used spell-checking tools, such as \emph{Grammarly}.

\end{document}